\documentclass[11pt]{article}
\usepackage[table]{xcolor}
\usepackage{acl}
\usepackage{times}
\usepackage{latexsym}

\usepackage{amsfonts}
\usepackage{amsmath}
\usepackage{graphicx,subfigure}
\usepackage{booktabs}
\usepackage{multirow}
\usepackage{url}
\usepackage{tabularx}
\usepackage{makecell}
\usepackage{multirow}
\usepackage{graphicx}
\usepackage{comment}
\usepackage{color}
\usepackage{kotex} 
\usepackage{hyperref}
\usepackage{multicol}
\usepackage{multirow}

\usepackage{stfloats}
\usepackage{placeins}
\usepackage{afterpage}
\usepackage{geometry}
\usepackage[table]{xcolor}
\hypersetup{
    colorlinks=true,
    linkcolor=blue,
    filecolor=magenta,      
}

\usepackage[T1]{fontenc}

\usepackage[utf8]{inputenc}

\usepackage{microtype}

\usepackage{inconsolata}

\usepackage{graphicx}
\usepackage{hyperref}
\title{UniSumEval: Towards Unified, Fine-Grained, Multi-Dimensional Summarization Evaluation for LLMs}

\author{Yuho Lee$^{1,}$\thanks{~~Equal Contribution.}, Taewon Yun$^{1,*}$,~ Jason Cai$^{2,}$\thanks{~~This work is conducted independently and is not related to the author(s)' position at Amazon.}~,~ Hang Su$^{2,\dagger}$,~ Hwanjun Song$^{1}$\thanks{~~Corresponding Author.}\\
$^{1}$Korea Advanced Institute of Science and Technology\\ $^{2}$AWS AI Labs\\
\ songhwanjun@kaist.ac.kr}

\newcommand{\algname}{\textsc{UniSumEval}}

\newcommand{\tobefilled}{{\textcolor{red}{\textbf{<to be filled>}}}} 

\definecolor{softpastelgreen}{RGB}{198, 233, 211}  

\newcolumntype{L}[1]{>{\raggedright\let\newline\\\arraybackslash\hspace{0pt}}m{#1}}
\newcolumntype{X}[1]{>{\centering\let\newline\\\arraybackslash\hspace{0pt}}p{#1}}
\newcolumntype{Y}[1]{>{\raggedleft\let\newline\\\arraybackslash\hspace{0pt}}m{#1}}

\begin{document}
 \maketitle
\begin{abstract}
Existing benchmarks for summarization quality evaluation often lack diverse input scenarios, focus on narrowly defined dimensions (\emph{e.g.}, faithfulness), and struggle with subjective and coarse-grained annotation schemes. To address these shortcomings, we create \algname{} benchmark, which extends the range of input context (\emph{e.g.}, domain, length) and provides fine-grained, multi-dimensional annotations. 
We use AI assistance in data creation, identifying potentially hallucinogenic input texts, and also helping human annotators reduce the difficulty of fine-grained annotation tasks.
With \algname{}, we benchmark nine latest language models as summarizers, offering insights into their performance across varying input contexts and evaluation dimensions. Furthermore, we conduct a thorough comparison of SOTA automated summary evaluators. Our benchmark data will be available at \url{https://github.com/DISL-Lab/UniSumEval-v1.0}. 

\end{abstract}

\section{Introduction}

Despite the enhanced quality of text summarization by large language models\,(LLMs), they still face persistent challenges like hallucination, information omission, and verbosity\,\cite{fabbri2022qafacteval, laban2023summedits}.
This multifaceted nature of text summaries inevitably demands manual evaluation by human experts, a labor-intensive and costly process. To streamline this evaluation process, recent efforts aim to design  \emph{human-like} automatic evaluators, 
such as {G-Eval}\,\cite{liu2023geval} and {FineSurE}\,\cite{song2024finesure},
which achieve a satisfactory correlation with human judgments.

Such evaluators are typically validated by examining their consistency with human judgments on established benchmark datasets, such as {FRANK} \cite{pagnoni2021understanding} and {TofuEval}\,\cite{tang2024tofueval}. Yet, these benchmark datasets have limitations in terms of input diversity, granularity of human annotations, and evaluation dimensions. 

Firstly, most existing benchmarks are restricted solely to a \emph{single domain}. The predominant focus is often on the news domain such as {SummEval} \cite{fabbri2021summeval} and {AggreFact} \cite{tang2023understanding}. This deficiency constrains the accurate evaluation of automated evaluators by failing to capture diverse input contexts across various domains.

Secondly, there is a lack of datasets that consider varying \emph{input types} and \emph{lengths} simultaneously. While these two factors have a significant impact on the summary quality, existing datasets are often limited to short, non-dialogue texts\,\cite{bhandari2020re, pagnoni2021understanding, laban2022summac}. 
Without considering these factors, they cannot adequately assess distinct perspectives across different input types and lengths. This includes correctly attributing statements to speakers in dialogue, preventing false information in articles with personally identifiable information\,(PII) redacted, and pinpointing key information in long texts.

Thirdly, no comprehensive datasets exist for \emph{fine-grained, multi-dimensional} summarization evaluation. Some benchmarks offer fine-grained annotations, such as fact verification at the sentence-level \cite{pagnoni2021understanding, laban2022summac, zhu2023annotating} and alignment at the key-fact\footnote{A key-fact refers a concise sentence conveying a single key piece of information, with at most 2-3 entities, also known as a semantic content unit\,\cite{bhandari2020re}.} level\,\cite{bhandari2020re, tang2024tofueval}, yet they suffer from either a limited evaluation dimension or coarse-grained human labels.

\begin{figure}[t!]
\begin{center}
\includegraphics[width=7.7cm]{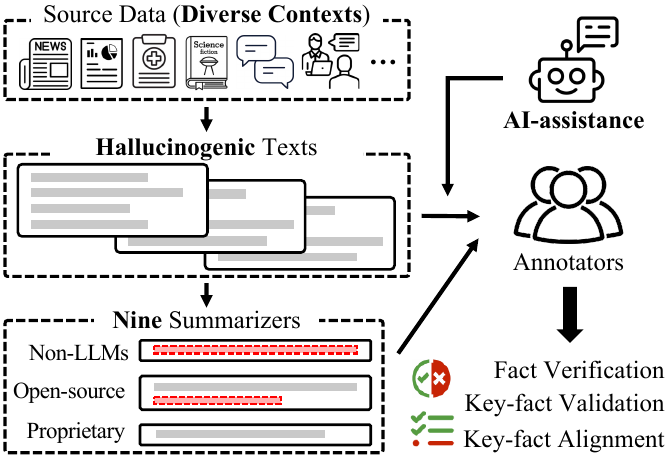}
\end{center}
\vspace*{-0.35cm}
\caption{\algname{} contains fine-grained and multi-dimensional human annotations with high IAA on various input domains, types, and lengths. We conduct AI-assisted manual evaluation on 2,025 hallucinogenic text-summary pairs with 2,509 human key-facts.}
\label{fig:overview}
\vspace*{-0.4cm}
\end{figure}

In this paper, we create {\algname{}} in Figure~\ref{fig:overview}, the first {one-size-fits-all} benchmark for fine-grained, multi-dimensional evaluation of automated evaluators. It includes: \textbf{Text Inputs} encompassing nine distinct domains (\emph{e.g.}, news, report, booking, meeting) that span from non-dialogue to dialogue, short texts to long texts containing up to 10,462 words, and even with redacted PII to simulate real-world scenarios; \textbf{Summaries} generated from nine latest summarizers across three categories, namely non-LLMs, open-source and proprietary LLMs; \textbf{Evaluation Dimension} covering three distinct evaluation aspects -- assessing faithfulness, information omission (completeness), and verbosity (conciseness) of the generated summaries; \textbf{AI-Assisted Manual Evaluation} collecting fine-grained, multi-dimensional human annotations with high IAA\footnote{We obtained the inter-annotator agreement\,(IAA) of 0.60 (Krippendorff’s $\alpha$ for fact verification, 0.88 (Gwet’s AC1) for key-fact validation, and 0.58 (Krippendorff’s $\alpha$) for key-fact alignment on average across {nine distinct domains}.} -- fact verification at the sentence level for faithfulness, key-fact validation and alignment of validated key-facts to each summary sentence for completeness and conciseness.

Unlike existing benchmarks, we identify \emph{hallucinogenic} texts, which can potentially trigger hallucination even for the latest LLMs. We then add these texts into our dataset, ensuring that our benchmark includes {challenging} text-summary pairs where the latest models actually generate and incur hallucinations. Also, the \emph{wide-ranging} text diversity within \algname{} enables a comprehensive evaluation of modern automated evaluators. This helps ascertain if the evaluators perform consistently across diverse input scenarios, an aspect overlooked in prior works. 

Using \algname{}'s fine-grained human labels, we further benchmark nine latest language models on summarization across \emph{multi-dimensional} aspects. We group them into three distinct categories: non-LLMs, open-source and proprietary LLMs. We then compare their performance across five key evaluation dimensions, including faithfulness, completeness, conciseness, abstractiveness, and domain stability. 
Next, we unveil the current progress of SOTA automated evaluators, including QA-, NLI-, and LLM-based methods, by comparing their evaluation scores with human judgments in \algname{}.

Our main contributions are as follows: (1) we are the first to create and release the fine-grained, multi-dimensional benchmark \algname{}, which covers diverse input contexts; (2) we develop an AI-assisted human evaluation protocol using Amazon Mechanical Turk\,(MTurk), achieving high IAA comparable to using expert linguistics, even for long input texts; (3) we systematically evaluate latest summarizers on faithfulness, completeness, conciseness, abstractiveness, and domain stability. The performance superiority among them varies depending on input domain, type, and length. PII redaction exacerbates the hallucination issue for all summarization models; (4) we conduct a thorough comparison of SOTA automated evaluators. Non-LLM evaluators perform poorly at verifying LLM-generated hallucinations. Evaluating conciseness (identifying unnecessary summary sentences) is harder than checking for faithfulness and completeness; (5) we release \algname{} to enable future research on automated evaluation.

\section{Related Work}\label{sec:rel}
\newcommand{\lightmidrule}{\specialrule{0.01em}{0.3em}{0em}}

\begin{table*}[t]
\scriptsize
\renewcommand{\arraystretch}{1.1}
\setlength{\tabcolsep}{6.8pt}
\begin{center}
\begin{tabular}{llllllll}
\toprule
\multicolumn{1}{c}{} &
  \multicolumn{2}{c}{Input Text Diversity} &
  \multicolumn{3}{c}{Human Annotation Scheme} &
  \multicolumn{2}{c}{Summary Generation} \\ 
\cmidrule(lr){2-3} \cmidrule(lr){4-6} \cmidrule(lr){7-8}
\multicolumn{1}{c}{} &
  \multicolumn{1}{c}{\makecell[c]{\# of Domains / \\ Input Type}} &
  \multicolumn{1}{c}{\makecell[c]{\# of Words\\ Avg. (Min -- Max)}} &
  \multicolumn{1}{c}{Granularity} &
  \multicolumn{1}{c}{Eval. Dim} &
  \multicolumn{1}{c}{\makecell[c]{Measurement}} &
  \multicolumn{1}{c}{\begin{tabular}[c]{@{}c@{}}Summaries \\      from LLMs\end{tabular}} &
  \multicolumn{1}{c}{Error Source} \\
\hline
\href{https://github.com/Yale-LILY/SummEval}{SummEval} &

  \cellcolor{pink}1 /   Non-dialogue &
  \cellcolor{pink}408 (106 -- 587) &
  \cellcolor{pink}Summary level &
  \cellcolor{softpastelgreen}Multiple &
  \cellcolor{pink}Likert-scale &
  \cellcolor{pink}No &
  \cellcolor{softpastelgreen}Realistic \\
\href{https://github.com/artidoro/frank}{FRANK} &

  \cellcolor{pink}1 /   Non-dialogue &
    \cellcolor{pink}528 (108 -- 1258) &
  \cellcolor{softpastelgreen}Sentence level &
  \cellcolor{pink}Single &
  \cellcolor{softpastelgreen}Percentage &
  \cellcolor{pink}No &
  \cellcolor{softpastelgreen}Realistic \\
\href{https://github.com/neulab/REALSumm}{REALSumm} &

  \cellcolor{pink}1 / Non-dialogue &
    \cellcolor{pink}745 (227 -- 1911) &
  \cellcolor{softpastelgreen}Key-fact level  &
  \cellcolor{pink}Single &
 \cellcolor{softpastelgreen}Percentage &
  \cellcolor{pink}No &
  \cellcolor{softpastelgreen}Realistic \\
\href{https://github.com/tingofurro/summac/}{SummaC} &

  \cellcolor{pink}1 /   Non-dialogue &
    \cellcolor{pink}583 (8 -- 11,667) &
  \cellcolor{pink}Summary level &
  \cellcolor{pink}Single &
  \cellcolor{pink}Binary-scale &
  \cellcolor{pink}No &
  \cellcolor{softpastelgreen}Realistic \\
\href{https://github.com/Liyan06/AggreFact}{AggreFACT} &

  \cellcolor{pink}1 /   Non-dialogue &
    \cellcolor{pink}496 (8 -- 11,667) &
  \cellcolor{pink}Summary level &
  \cellcolor{pink}Single &
    \cellcolor{pink}Binary-scale &
  \cellcolor{pink}No &
   \cellcolor{softpastelgreen}Realistic\\
\href{https://github.com/martiansideofthemoon/longeval-summarization}{LongEval} &

  \cellcolor{pink}2 /   Non-dialogue  &
    \cellcolor{softpastelgreen}4,917 (1,009 -- 12,319) &
   \cellcolor{softpastelgreen}Key-fact level &
  \cellcolor{pink}Single &
  \cellcolor{softpastelgreen}Percentage &
  \cellcolor{pink}No &
   \cellcolor{softpastelgreen}Realistic \\
\href{https://github.com/kite99520/DialSummEval}{DialSummEval} &

  \cellcolor{pink}1 / Dialogue &
    \cellcolor{pink}130 (24 -- 488) &
  \cellcolor{pink}Summary level &
  \cellcolor{softpastelgreen}Multiple &
  \cellcolor{pink}Likert-scale &
  \cellcolor{pink}No &
  \cellcolor{softpastelgreen}Realistic \\
\href{https://github.com/731935354/Dia-Sum-Fact}{DiaSumFact} &

  \cellcolor{pink}2 / Dialogue &
    \cellcolor{pink}247 (24 -- 585) &
  \cellcolor{softpastelgreen}Sentence level &
  \cellcolor{pink}Single &
  \cellcolor{softpastelgreen}Percentage &
  \cellcolor{pink}No &
  \cellcolor{softpastelgreen}Realistic \\
\href{https://github.com/amazon-science/tofueval}{TofuEval} &

  \cellcolor{pink}2 / Dialogue &
    \cellcolor{softpastelgreen}950 (710 -- 1,199) &
  \cellcolor{pink}Mixed level   &
  \cellcolor{softpastelgreen}Multiple &
  \cellcolor{pink}Mixed-scale &
  \cellcolor{softpastelgreen}Yes &
  \cellcolor{softpastelgreen}Realistic \\ 
\href{https://github.com/salesforce/factualNLG}{SummEdits} &

  \cellcolor{softpastelgreen}9 / Mixed &
    \cellcolor{pink}705 (39 -- 2,569) &
  \cellcolor{pink}Summary level  &
  \cellcolor{pink}Single &
 \cellcolor{pink}Ternary-scale &
  \cellcolor{softpastelgreen}Yes &
  \cellcolor{pink}Synthetic \\
\hline
\textbf{UniSumEval} &

  \cellcolor{softpastelgreen}\textbf{9 / Mixed} &
    \cellcolor{softpastelgreen}\textbf{2,092 (21 -- 10,462)} &
  \cellcolor{softpastelgreen}\textbf{Sentence \& Key-fact} &
  \cellcolor{softpastelgreen}\textbf{Multiple} &
  \cellcolor{softpastelgreen}\textbf{Percentage} &
  \cellcolor{softpastelgreen}\textbf{Yes} &
  \cellcolor{softpastelgreen}\textbf{Realistic} \\
\Xhline{0.75pt}
\end{tabular}
\vspace*{-0.3cm}
\caption{Comparison of \algname{} with the ten existing summarization evaluation benchmarks. The mixed level of granularity indicates that the evaluation dimensions have annotations at either the sentence or summary level. The mixed level of measurement indicates that a different scale is used for each dimension.
}
\label{tab:existing_benchmarks}
\end{center}
\vspace*{-0.55cm}
\end{table*}

\noindent\textbf{Evaluation Benchmarks.}
Existing benchmarks in summary quality evaluation have predominantly concentrated on assessing the performance of automated metrics in evaluating faithfulness\,\cite{bhandari2020re, pagnoni2021understanding, laban2022summac, tang2023understanding}. These benchmarks have generally been limited to short and non-dialogue texts. This limitation has spurred the development of new benchmarks that specifically address either longer texts\,\cite{krishna2023longeval} or dialogue-based texts\,\cite{gao2022dialsummeval, zhu2023annotating, laban2023summedits, tang2024tofueval}. Additionally, another segment of benchmarks has expanded beyond the dimension of faithfulness to include relevance and coherence\,\cite{fabbri2021summeval, gao2022dialsummeval, tang2024tofueval}, also enhancing the granularity of annotations\,\cite{zhu2023annotating}. 
Recently, more advanced benchmarks have been developed, utilizing the power of LLMs. SummEdits\,\cite{laban2023summedits} expands seed summaries to non-factual ones by synthetic editing using LLMs but focuses solely on faithfulness.
TofuEval\,\cite{tang2024tofueval} generates topic-based summaries using LLMs but limits its scope only to dialogues.

\smallskip\smallskip
\noindent\textbf{Automated Evaluation.}
Conventional similarity-based metrics, such as ROUGE-1/2/L \cite{lin2004rouge}, BERTScore \cite{zhang2019bertscore}, and BARTScore \cite{yuan2021bartscore} have shown poor correlation with human judgments.
In response, natural language inference\,(NLI)-based methods have emerged to verify the faithfulness of summaries by retrieving relevant evidence in their input texts\,\cite{laban2022summac, tang2024minicheck, zha2023alignscore}. Similarly, Question Answering\,(QA)-based methods involve generating relevant questions from the reference text and answering them based on the generated content\,\cite{fabbri2022qafacteval, zhong2022towards}. While both directions have shown improved performance, they are generally limited to faithfulness evaluation and also require training specialized models.
Recently, LLM-based evaluators have been proposed as reference-free, automated evaluators usable in various contexts \cite{liu2023geval, song2024finesure}. 
While they show promise with short news articles, they still struggle with fine-grained evaluations, and their performance across various domains and input types has not been properly investigated.

\section{\algname{} Pipeline} \label{sec:method}

Our data creation pipeline consists of four consecutive steps in the following sections.
Table~\ref{tab:existing_benchmarks} contrasts \algname{} with existing benchmarks across various aspects, including input diversity, annotation schemes, and data generation. The statistics of \algname{} are provided in Appendix~\ref{app:datasets}.

\subsection{Input Text Sourcing}\label{subsec:subsec3.1}

We use nine source datasets to construct our benchmark dataset: {Wikihow} (lifestyle)\,\cite{koupaee2018wikihow}, {CNN/DM} (news)\,\cite{nallapati2016abstractive}, {GovReport} (report)\,\cite{huang2021efficient}, {PubMed} (medical literature)\,\cite{cohan2018discourse}, {SQuALITY} (science fiction)\,\,\cite{wang2022squality}, {MultiWOZ} (booking conversation)\,\cite{zang2020multiwoz}, {DialogSum} (daily life conversation)\,\cite{chen2021dialogsum}, {MediaSum} (interview)\,\cite{zhu2021mediasum}, and {MeetingBank} (meeting) \cite{hu2023meetingbank}.
This selection ensures that each source dataset covers nine distinct domains and maintains a balanced distribution of text types (dialogue, non-dialogue) and lengths (short text, long text).

\subsection{Summary Generation and Selection}\label{subsec:subsec3.2}

\noindent\textbf{Summary Generation.} 
We randomly sample 200 input texts from the test set of each source dataset. Then, we generate summaries using the nine latest language models as summarizers, chosen for their widespread usage. These models are classified into three categories: \emph{non-LLMs}, including fine-tuned BART-Large \cite{lewis2020bart} and Flan-T5-Large \cite{chung2024scaling}, each with fewer than 700M parameters; \emph{open-source LLMs}, including Phi-2 \cite{javaheripi2023phi}, Llama2-13B-Chat \cite{touvron2023llama}, instruction-tuned Mistral-7B \cite{jiang2023mistral} and Mixtral-8x7B \cite{jiang2024mixtral}; and \emph{proprietary LLMs}, including GPT-3.5-turbo, GPT-4-turbo \cite{achiam2023gpt}, and Claude2.1. See Appendix~\ref{app:summary_generation_and_prompts} for model details and prompts used to generate summaries.


\smallskip\smallskip
\noindent\textbf{Hallucinogenic Text Selection.} The latest models, such as Claude2.1 and GPT-4-turbo, generate hallucinations that are subtle and less common. Hence, to create a more challenging benchmark, we identify \emph{hallucinogenic} texts, which have the potential to induce hallucination even with the latest models. Specifically, an input text is classified as hallucinogenic if at least one of the nine models we used generates a hallucination from it. 
We perform an LLM-based automatic evaluation of all text-summary pairs to generate sentence-level binary labels for faithfulness, accompanied by a one-sentence rationale for each label (see Appendix~\ref{app:ai_assistant_prompt} for details including the prompt).
Based on the LLM-based automatic evaluation, we re-sample 25 hallucinogenic texts from the 200 sampled texts for each source. As a result, we finally obtain a total of 2,025 text-summary pairs (= 9 datasets $\times$ 25 hallucinogenic texts $\times$ 9 summarizers).

\subsection{Fine-Grained Annotation Tasks} 
\label{subsec:subsec3.3}

\footnotetext[3]{
We generate initial key-facts using GPT-4-turbo with a tuned prompt. These key-facts are then reduced using other LLMs, including GPT-3.5-turbo, Claude2.1, and Mixtral-8x7B, by eliminating those the majority disagree with. This results in an average of 8 and 14 potential key-facts for short and long texts, respectively. See Appendix~\ref{app:key-fact_extraction_prompt} for the prompts for the key-fact generation and reduction with multiple LLMs.}

We collect fine-grained human labels for multi-dimensional aspects of summary evaluation. {The conventional dimensions like coherence and relevance is not adequate for fine-grained evaluation, due to the ambiguity in their definitions. Thus, we follow the three fine-grained dimensions suggested in the recent work \cite{song2024finesure}, namely faithfulness, completeness, and conciseness.} 

Faithfulness is assessed at the sentence level by \emph{fact verification}, a task of assigning a binary label (Yes/No) indicating whether a sentence has factual errors across four predefined categories. These include Out-of-Article Error as an extrinsic error, and Entity Error, Sentence Error, Relation Error as the three subcategories of intrinsic errors {(see Appendix~\ref{app:factual_error_types} for more details on the error taxonomy)}. If the response is Yes, we then ask the respondents to identify error types using a multichoice form.

\begin{figure}[t]
\begin{center}
\includegraphics[width=7.7cm]{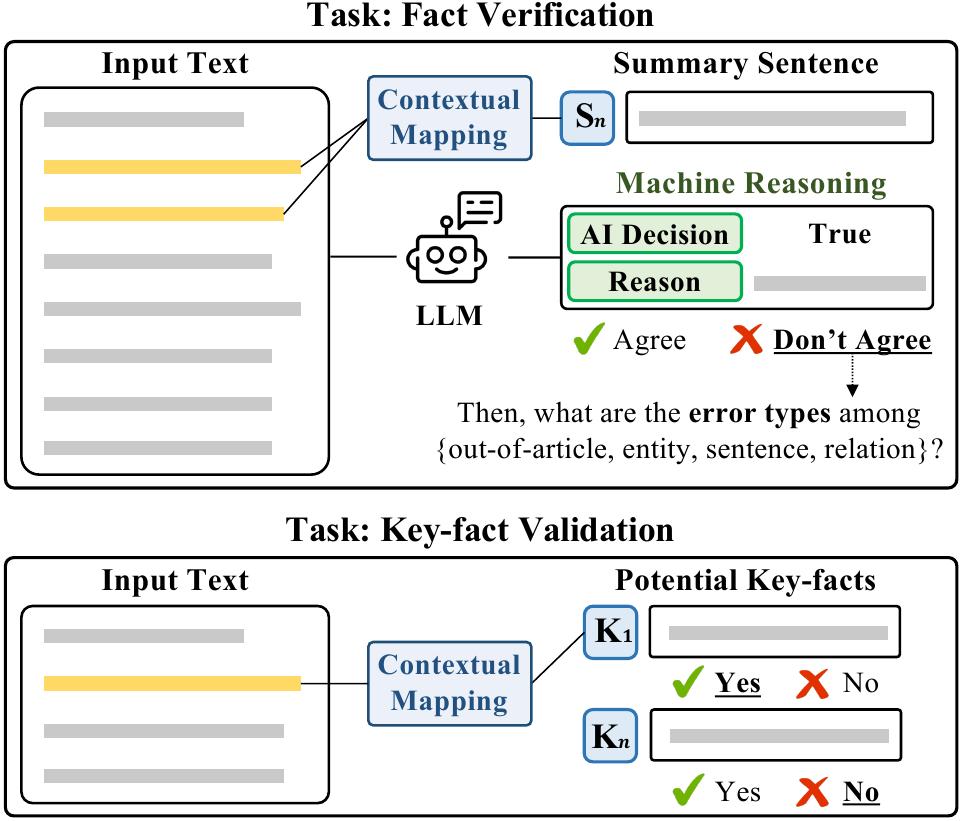}
\end{center}
\vspace*{-0.45cm}
\caption{AI-assisted fine-grained manual evaluation. }
\label{fig:ai_assist_framework}
\vspace*{-0.45cm}
\end{figure}

In contrast, completeness and conciseness are annotated at the key-fact level using two different tasks. More specifically, we carefully generate the list of {potential key-facts} using multiple LLMs\footnotemark[3] and then conduct \emph{key-fact validation}, a human annotation task for verifying if the information in each generated key-fact is significant, factually correct, and relevant with respect to its source text.
It enables the identification of human verified key-facts from the generated ones. 
%
Next, we perform another annotation task of \emph{key-fact alignment}, matching each human key-fact to all summary sentences from which they can be inferred. \looseness=-1

Across the three annotation tasks, we can compute the percentage\,(\%) scores for three evaluation dimensions at the summary level: (1) \emph{faithfulness}, the proportion of factually correct summary sentences; (2) \emph{completeness}, the proportion of key-facts inferable from the summary; and (3) \emph{conciseness}, the proportion of summary sentences aligned with the key-facts.
%
See Appendix \ref{app:summary_performance_calculation} for the detailed formulas.

\subsection{AI-Assisted Manual Evaluation}\label{subsec:subsec3.4}

Unlike key-fact alignment, which only requires annotators to align key-facts and summary sentences, fact verification and key-fact validation tasks necessitate a thorough understanding of the input text. This issue evidently leads to a significant drop in IAA for manual evaluation when input texts are lengthy, as humans struggle to manage large volumes of information simultaneously\,\cite{krishna2023longeval}.
Hence, we devise AI-assisted manual evaluation in Figure \ref{fig:ai_assist_framework}, which helps achieve high IAA among non-expert human annotators for long texts.

We apply \emph{contextual mapping}, a task of highlighting sentences in an input text that rank in the top 30\% for similarity\footnote{We use the pre-trained BERT\,\cite{kenton2019bert} to compute the similarity. The choice of a 30\% threshold is intentionally set as a conservative value, ensuring high recall to encompass all relevant input sentences.} with a target summary sentence in fact verification (or a key-fact in key-fact validation) being evaluated. This aids annotators by providing contextual cues and narrowing down the relevant sections of the input text.
Additionally, for fact verification, we plug in \emph{machine}\emph{-based reasoning}, the assistance of providing the decision by an automatic evaluator. This is based on the inferred sentence-level factuality labels and reasoning obtained in the hallucinogenic text selection. To identify annotators who blindly endorse AI decisions, we replace 20\% of the tasks with attention checks, where the correct answers are predetermined. It helps us detect unfaithful responses based on annotators' (dis)agreement with the AI decision. See Appendix \ref{sec:attention} for details on the attention check.

This systematic integration of AI into manual evaluation enables not only to achieve high IAA even for long input texts but also to promote a more cost-effective assessment.

\subsection{Annotation Procedure}
We conduct AI-assisted manual evaluation using MTruk on our three tasks. 
We select annotators who pass an English qualification test, with an approval rating above 95\% and at least 1,000 accepted HITs. We collect human annotations from three independent qualified annotators for 8,133 summary sentences in fact verification, and for 2,673 key-facts in key-fact validation. The 2,673 potential key-facts are reduced to 2,509 human key-facts. Consequently, annotations for key-fact alignment are collected for all possible pairs between human key-facts and summary sentences of the same input text, \emph{i.e.}, 101,013 pairs in total.
Annotators are paid 50\% above the U.S minimum wage and receive \$25 bonuses for every 500 HITs.
See Appendix~\ref{app:annotationdetail} for more details on our annotation tasks.

\begin{table*}[!ht]
\begin{center}
\scriptsize

\setlength{\tabcolsep}{2.8pt}
\renewcommand{\arraystretch}{1}
\begin{tabular}{ccccccccccccc}
\toprule
\multirow{3}{*}{\makecell[c]Evaluation Dim.} & 
\multirow{3}{*}{Annotation Task} &
\multirow{3}{*}{Granularity} &
\multicolumn{4}{c}{Short} & 
\multicolumn{5}{c}{Long} & 
\multirow{2}{*}{Avg. IAA} \\
\cmidrule(lr){4-7} \cmidrule(lr){8-12}
 &
 &  
 & News 
 & LifeStyle 
 & Booking 
 & Daily Life 
 & Report 
 & Med Lit 
 & Sci-fi 
 & Interview 
 & Meeting &  \\
\midrule

Faithfulness & Fact Verification
& Sentence level & 0.63  & 0.78  & 0.66 &  0.78 & 0.49  & 0.42 & 0.48 & 0.43 & 0.65 & 0.60                               \\
\midrule

\multirow{2}{*}{\makecell{Completeness \& \\ Conciseness}} & Key-fact Validation & Key-fact level & \makecell{0.90} 
& \makecell{0.96} 
& \makecell{0.95} &\makecell{0.91} 
& \makecell{0.98} 
& \makecell{0.96} 
& \makecell{0.67} 
& \makecell{0.84} 
& \makecell{0.77} 
& \makecell{0.88} \\

 &Key-fact Alignment
& Key-fact level & 0.61  & 0.80  & 0.69 &  0.71 & 0.43  & 0.41 & 0.47 & 0.54 & 0.56 & 0.58                               \\
\bottomrule
\end{tabular}
\vspace*{-0.2cm}
\caption{IAA scores of human labels in \algname{} for each data domain. The subset is categorized as "short" if the average number of words is less than 900, otherwise "long". }
\label{tab:totalIAA}
\end{center}
\vspace*{-0.6cm}
\end{table*}

\section{\algname{} Quality Assessment}\label{sec:exp-annotation}

\noindent\textbf{Evaluation Metric.} 
We report IAA for three fine-grained annotation tasks in \algname{}: fact verification, key-fact validation and alignment. We use Krippendorff's $\alpha$ \cite{krippendorff2011computing} by default, but for key-fact validation, where there is significant label imbalance, we use Gwet's AC1 \cite{wongpakaran2013comparison} due to its enhanced robustness.\looseness=-1

\subsection{Inter-Annotator Agreement}

\begin{table}[t!]
\vspace*{-0.0cm}
\centering
\footnotesize
\begin{tabular}{ccc}
\toprule
Benchmark & Annotation Task & IAA \\
\midrule
FRANK & Sentence level fact verification & 0.63 \\
DiaSumFact$^{\dagger}$ & Sentence level fact verification  & 0.49 \\
TofuEval & Sentence level fact verification  & 0.40 \\
LongEval$^{\dagger}$ & Key-fact level fact verification & 0.64 \\
REALSumm & Key-fact alignment & 0.66 \\
\midrule
\multirow{2}{*}{\textbf{\algname{}}} & Sentence level fact verification  & 0.60 \\
 & {Key-fact alignment} & 0.58\\
\bottomrule
\end{tabular}
\vspace*{-0.2cm}
\caption{IAA comparison across the existing benchmarks with fine-grained labels. We report Krppendorff's $\alpha$ by default. ${\dagger}$: we copy the Fleiss' $\kappa$ value in the original paper due to the absence of annotator-level labels.}
\vspace*{-0.45cm}
\label{tab:IAA_comparison}
\end{table}

\noindent\textbf{Overall Assessment.}
Table \ref{tab:totalIAA} shows the IAA scores of \algname{} across nine data domains.
In Table~\ref{tab:IAA_comparison}, we compare \algname{} with existing benchmarks annotated with fine-grained labels. \textbf{\algname{} stands out as the only benchmark supporting multi-dimensional evaluation of automated evaluators}, while others only focus on either faithfulness (via fact verification) or completeness (via key-fact alignment). Additionally, our benchmark exhibits IAA better than or comparable to others, even with the comprehensive inclusion of varying input contexts, as evidenced by Table \ref{tab:existing_benchmarks}. 
Particularly, TofuEval got an IAA of 0.40 with linguistic experts in the meeting domain (long text), while we get a higher IAA of 0.65 in the same domain even with non-expert labels through AI-assisted manual evaluation (see Table \ref{tab:totalIAA}). 

\smallskip\smallskip
\noindent\textbf{Impact of Input Context.} The \algname{}'s input diversity enables a thorough analysis of how IAA varies across different domains in manual evaluation. In Table \ref{tab:totalIAA}, we note a high Pearson correlation of 0.89 in IAA between fact verification and key-fact alignment tasks across various data domains. Also, input characteristics, such as long-form texts and professional texts including reports, medical literature, science fiction, and interviews, have a negative impact on IAA.
Therefore, \textbf{input contexts significantly influence the IAA score of fine-grained human annotation tasks.}

\smallskip\smallskip
\noindent\textbf{Efficacy of AI Assistance.}
We use LLM-based evaluation for two purposes -- selecting hallucinogenic input texts; and assisting annotators in fact verification. For the former, 93.3\% of the selected input texts are confirmed to produce real hallucinations in at least one summarizer. 
For the latter, we conduct an ablation study on the fact verification task with three variants: (1) highlighting relevant input sentences through contextual mapping; (2) providing factuality labels estimated by the LLM; and (3) providing the labels with reasoning by the LLM. \textbf{The IAA for fact verification improves significantly with AI assistance} from 0.28 to 0.55 by adding (2); and further to 0.63 by adding (3).

In particular, we highlight that human annotators do not indiscriminately accept machine labels. They reveal that 19.31\% of the factuality errors flagged by the LLM are inaccurately identified, while 2.59\% of sentences deemed error-free by the LLM actually contains factual inaccuracy. See Appendix~\ref{app:IAA_human_machine_mismatch} for further analysis.

\section{Benchmarking Summarizers}
\label{sec:exp-summarizer}

\begin{figure*}[t!]
\begin{center}
\vspace*{-0.35cm}
\includegraphics[width=15.5cm]{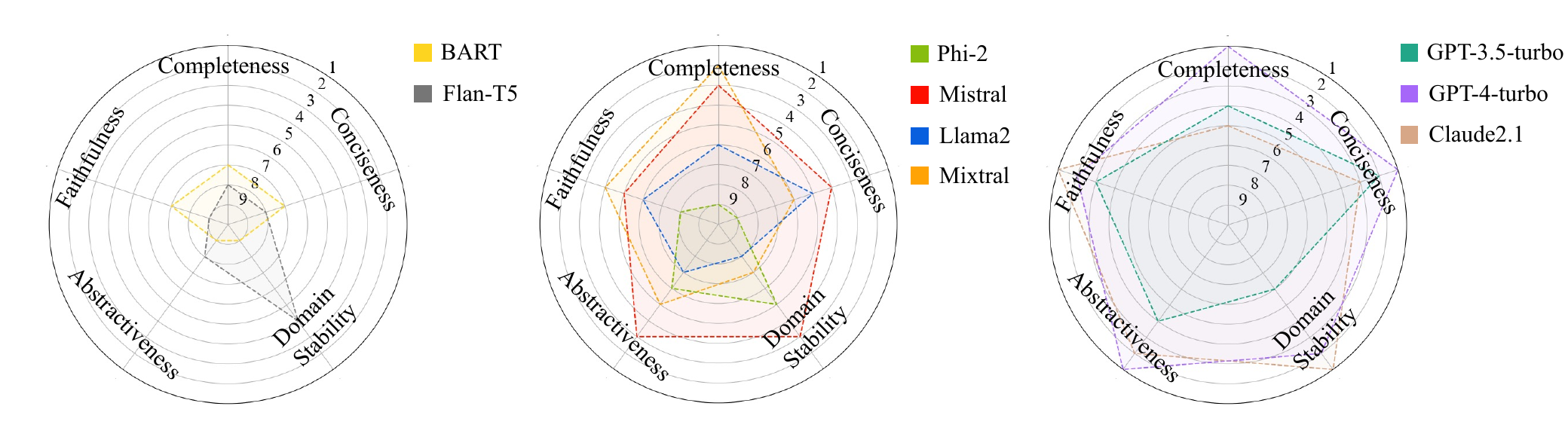}
\end{center}
\vspace*{-0.65cm}
{\small \hspace*{1.37cm} (a) Non-LLMs. \hspace*{2.3cm} (b) Open-source LLMs. \hspace*{1.9cm} (c) Proprietary LLMs.}
\vspace*{-0.2cm}
\caption{Performance ranking (1-9) of the nine recent summarizers across the five evaluation dimensions. The summarizers are categorized into three distinct groups: non-LLMs, open-source LLMs, and proprietary LLMs.
}
\label{fig:radarchart}
\vspace*{-0.1cm}
\end{figure*}


\definecolor{firstgreen}{HTML}{7CCD70}
\definecolor{secondgreen}{HTML}{A8E6A3}
\definecolor{green1}{HTML}{b2e79f}

\definecolor{firstorange}{HTML}{91DB72}
\definecolor{secondorange}{HTML}{228B22}
\definecolor{orange1}{HTML}{FBD783}

\definecolor{firstred}{HTML}{228B22}
\definecolor{secondred}{HTML}{91DB72}
\definecolor{red1}{HTML}{FAA0A0}
\definecolor{bettergreen}{HTML}{228B22}

\begin{table*}[t]
\begin{center}
\scriptsize
\setlength{\tabcolsep}{4.7pt}
\renewcommand{\arraystretch}{1.05}
\begin{tabular}{ccccccccccccccccc}
\toprule
\multirow{3}{*}{\makecell[c]{Model \\ Type}} &
  \multirow{3}{*}{\makecell[c]{Summ. \\ Model}} &
  \multicolumn{6}{c}{Non-Dialogue} &
  \multicolumn{6}{c}{Dialogue} &
  \multicolumn{3}{c}{\multirow{2}{*}{\makecell[c]{Avg. Score \\ (Model-wise) \\ }}} \\
\cmidrule(lr){3-8} \cmidrule(lr){9-14}
 &
   &
  \multicolumn{3}{c}{Short} &
  \multicolumn{3}{c}{Long} &
  \multicolumn{3}{c}{Short} &
  \multicolumn{3}{c}{Long} &
  \multicolumn{3}{c}{} \\
\cmidrule(lr){3-5} \cmidrule(lr){6-8} \cmidrule(lr){9-11} \cmidrule(lr){12-14} \cmidrule(lr){15-17} 
 &
   &
  Faith &
  Comp &
  Conc &
  Faith &
  Comp &
  Conc &
  Faith &
  Comp &
  Conc &
  Faith &
  Comp &
  Conc &
  Faith &
  Comp &
  Conc \\
\midrule
\multirow{2}{*}{\makecell[c]{Non \\ LLM}} &
  BART\textsubscript{large} &
  \cellcolor{red1}87.6 &
  \cellcolor{orange1}67.9 &
  \cellcolor{red1}86.6 &
  \cellcolor{red1}89.5 &
  \cellcolor{red1}16.3 &
  \cellcolor{red1}60.1 &
  \cellcolor{red1}81.5 &
  \cellcolor{orange1}47.8 &
  \cellcolor{orange1}79.6 &
  \cellcolor{red1}77.9 &
  \cellcolor{red1}30.9 &
  \cellcolor{red1}72.2 &
  \cellcolor{red1}84.1 &
  \cellcolor{red1}40.7 &
  \cellcolor{red1}74.6 
  \\ 
 &
  Flan-T5\textsubscript{large} &
  \cellcolor{red1}90.4 &
  \cellcolor{red1}43.5 &
  \cellcolor{red1}83.8 &
  \cellcolor{red1}83.3 &
  \cellcolor{red1}16.3 &
  \cellcolor{orange1}69.0 &
  \cellcolor{red1}72.3 &
  \cellcolor{red1}44.5 &
  \cellcolor{red1}76.3 &
  \cellcolor{red1}75.4 &
  \cellcolor{red1}32.2 &
  \cellcolor{red1}67.4 &
  \cellcolor{red1}80.4 &
  \cellcolor{red1}34.1 &
  \cellcolor{red1}74.1 
  \\
\hline
\multirow{4}{*}{\makecell[c]{Open \\ Source \\ LLM}}
&
  Phi-2 &
  \cellcolor{red1}72.3 &
  \cellcolor{red1}56.6 &
  \cellcolor{red1}71.1 &
  \cellcolor{red1}84.5 &
  \cellcolor{red1}16.9 &
  \cellcolor{red1}43.6 &
  \cellcolor{red1}86.8 &
  \cellcolor{red1}36.9 &
  \cellcolor{red1}60.8 &
  \cellcolor{red1}81.2 &
  \cellcolor{red1}25.6 &
  \cellcolor{red1}49.5 &
  \cellcolor{red1}81.2 &
  \cellcolor{red1}34.0 &
  \cellcolor{red1}56.3 
  \\
&
  Mistral\textsubscript{7B-Inst} &
  \cellcolor{orange1}95.9 &
  \cellcolor{green1}77.7 &
  \cellcolor{orange1}89.3 &
  \cellcolor{orange1}96.9 &
  \cellcolor{green1}43.1 &
  \cellcolor{orange1}70.4 &
  \cellcolor{orange1}94.7 &
  \cellcolor{green1}66.2 &
  \cellcolor{orange1}79.6 &
  \cellcolor{orange1}92.9 &
  \cellcolor{green1}59.4 &
  \cellcolor{orange1}75.1 &
  \cellcolor{orange1}95.1 &
  \cellcolor{green1}61.6 &
  \cellcolor{orange1}78.6 \\
 &
  Llama2\textsubscript{13B-Chat} &
  \cellcolor{orange1}96.3 &
  \cellcolor{red1}61.6 &
  \cellcolor{green1}92.6 &
  \cellcolor{orange1}92.4 &
  \cellcolor{orange1}29.1 &
  \cellcolor{red1}63.6 &
  \cellcolor{orange1}91.7 &
  \cellcolor{red1}40.3 &
  \cellcolor{green1}79.9 &
  \cellcolor{orange1}86.7 &
  \cellcolor{orange1}35.8 &
  \cellcolor{orange1}73.1 &
  \cellcolor{orange1}91.8 &
  \cellcolor{orange1}41.7 &
  \cellcolor{orange1}77.3 \\
 &
  Mixtral\textsubscript{8x7B-Inst} &
  \cellcolor{green1}97.7 &
  \cellcolor{green1}80.9 &
  \cellcolor{orange1}87.7 &
  \cellcolor{orange1}97.0 &
  \cellcolor{green1}44.2 &
  \cellcolor{orange1}63.6 &
  \cellcolor{orange1}96.1 &
  \cellcolor{green1}68.4 &
  \cellcolor{orange1}77.1 &
  \cellcolor{green1}95.6 &
  \cellcolor{green1}60.5 &
  \cellcolor{orange1}74.0 &
  \cellcolor{green1}96.6 &
  \cellcolor{green1}63.5 &
  \cellcolor{orange1}75.6 \\
\hline
\multirow{3}{*}{\makecell[c]{Prop. \\ LLM}} &
  GPT-3.5\textsubscript{turbo} &
  \cellcolor{orange1}95.5 &
  \cellcolor{orange1}71.7 &
  \cellcolor{green1}91.5 &
  \cellcolor{green1}97.2 &
  \cellcolor{orange1}42.4 &
  \cellcolor{green1}81.5 &
  \cellcolor{green1}98.5 &
  \cellcolor{orange1}62.8 &
  \cellcolor{green1}83.2 &
  \cellcolor{green1}93.8 &
  \cellcolor{orange1}49.3 &
  \cellcolor{green1}79.2 &
  \cellcolor{orange1}96.2 &
  \cellcolor{orange1}56.6 &
  \cellcolor{green1}83.8 \\
 &
  GPT-4\textsubscript{turbo} &
  \cellcolor{green1}97.3 &
  \cellcolor{green1}79.3 &
  \cellcolor{green1}90.5 &
  \cellcolor{green1}98.1 &
  \cellcolor{green1}45.0 &
  \cellcolor{green1}82.6 &
  \cellcolor{green1}99.0 &
  \cellcolor{green1}76.3 &
  \cellcolor{green1}86.9 &
  \cellcolor{orange1}92.6 &
  \cellcolor{green1}64.5 &
  \cellcolor{green1}81.9 &
  \cellcolor{green1}96.8 &
  \cellcolor{green1}66.3 &
  \cellcolor{green1}85.5 \\
 &
  Claude2.1 &
  \cellcolor{green1}96.3 &
  \cellcolor{orange1}64.2 &
  \cellcolor{orange1}84.6 &
  \cellcolor{green1}99.3 &
  \cellcolor{orange1}32.0 &
  \cellcolor{green1}74.7 &
  \cellcolor{green1}98.4 &
  \cellcolor{orange1}60.4 &
  \cellcolor{red1}76.0 &
  \cellcolor{green1}95.5 &
  \cellcolor{orange1}44.8 &
  \cellcolor{green1}85.8 &
  \cellcolor{green1}97.4 &
  \cellcolor{orange1}50.4 &
  \cellcolor{green1}80.3 \\ 
\midrule
\multicolumn{2}{c}{\makecell[c]{Avg. Score (Context-wise) \\ }} &
  \multicolumn{1}{|c}{92.1} & 
  67.0 & 86.4 &
  \multicolumn{1}{|c}{93.1} &
  31.7 & 67.7 &
  \multicolumn{1}{|c}{91.0} &
  56.0 & 77.7 &
  \multicolumn{1}{|c}{88.0} &
  44.8 & 73.1 &
  \multicolumn{1}{|c}{91.1} &
  50.4 & 76.2 \\
\bottomrule
\end{tabular}
\vspace*{-0.2cm}
\caption{
Faithfulness (Faith), completeness (Comp), and conciseness (Cons) scores (\%) based on human annotations across different input contexts. \textcolor{bettergreen}{Green}, \textcolor{orange}{orange} and \textcolor{red}{red} indicate the performance ranking intervals of the model for each dimension, corresponding to top (rank 1-3), middle (rank 4-6), and bottom (rank 7-9) tiers, respectively.}
\label{tab:overall_score}
\end{center}
\vspace*{-0.6cm}
\end{table*}

\noindent\textbf{Evaluation Dimension.} We evaluate the nine summarizers across five crucial evaluation dimensions for text summarization. In addition to \emph{faithfulness}, \emph{completeness}, and \emph{conciseness}, we add two more dimensions: \emph{domain stability}, the consistency of a summarizer's performance across the nine domains; \emph{abstractiveness}, the extent to which a summary generates novel sentences or phrases, leading to a more coherent and condensed summary.

\smallskip\smallskip
\noindent\textbf{Evaluation Metric.} We report percentage scores (in Section \ref{subsec:subsec3.3}) of faithfulness, completeness, and conciseness, computed by using fine-grained human annotations. For domain stability, we calculate the average of the three percentage scores to obtain a composite score, and then measure domain inconsistency by computing the gap between the highest and lowest composite ones. For abstractiveness, we use the average of novel 1/3/5-grams following \citet{song2023enhancing}. See Appendix~\ref{app:summary_performance_calculation} for their detailed definitions.

\vspace*{-0.1cm}
\subsection{Comparison over Nine Summarizers}
\vspace*{-0.1cm}

Figure~\ref{fig:radarchart} shows the overall performance rankings aggregated across the all text domains, types, and lengths in \algname{}. The proprietary LLMs notably outperform the non-LLMs and open-source LLMs across the all aspects. GPT-4-turbo is the best summarizer in general, while Claude2.1 has the best faithfulness and domain stability. Proprietary LLMs also exhibit an interesting behavior -- \textbf{no statistical relationship between faithfulness and abstractiveness} -- contradicting recent findings\,\cite{maynez2020faithfulness, ladhak2022faithful} that summaries with higher abstractiveness are more prone to trigger hallucination. Statistical analysis on this can be found in Appendix~\ref{app:abstractive_faithfulness}.

\subsection{Impact of Domain, Type, and Length}

Table~\ref{tab:overall_score} breaks down the performance in Figure \ref{fig:radarchart}, highlighting the impact of input contexts -- input domain, type, and length -- on each summarizer.

Firstly, \textbf{the superiority in summarization quality among the summarizers varies depending on input domain, type, and length.} The general tendency among the three summarizer categories is consistent, while within each category, there are considerable changes in the summarizers' rankings (see the color changes for each column). 

Secondly, \textbf{most summarizers experience a significant performance drop in completeness and conciseness with lengthy input texts.} Particularly, such a drop is more noticeable when dealing with non-dialogue than dialogue. This confirms that identifying key-facts in summary generation is more challenging with longer input texts. 

\begin{table}[!t]
\begin{center}
\footnotesize
\renewcommand{\arraystretch}{1.1}
\setlength{\tabcolsep}{7.5pt}
\begin{tabular}{cccc}
\toprule
Type & Model & Unredaction & Redaction \\ 
\midrule
\multirow{2}{*}{\makecell{Non \\ LLM}}     
& BART\textsubscript{large}        & 82.8  & 79.0 \textcolor{red}{(-3.8)}  \\
& Flan-T5\textsubscript{large}     & 78.6  & 75.7 \textcolor{red}{(-2.9)} \\
\midrule
\multirow{2}{*}{\makecell{Open \\ Source }}
& Mistral\textsubscript{7B-Inst}   & 97.9  & 95.7 \textcolor{red}{(-2.2)}  \\
& Mixtral\textsubscript{8x7B-Inst} & 93.9  & 92.6 \textcolor{red}{(-1.3)} \\
\midrule
\multirow{2}{*}{\makecell{Prop. \\ LLM}}  
& GPT-3.5\textsubscript{turbo}     & 98.0  & 97.0 \textcolor{red}{(-1.0)} \\
& GPT-4\textsubscript{turbo}            & 100.0 & 98.0 \textcolor{red}{(-2.0)}\\
\bottomrule
\end{tabular}
\vspace*{-0.3cm}
\caption{Impact of PII redaction on faithfulness on booking conversation (source: MultiWOZ).} 
\label{tab:redaction_results_abc}
\end{center}
\vspace*{-0.7cm}
\end{table}

\begin{table*}[t]
\begin{center}
\scriptsize
\setlength{\tabcolsep}{8pt}
\renewcommand{\arraystretch}{1.1}
\begin{tabular}{ccccccccccc}
\toprule
\multirow{3}{*}{\makecell{Model \\ Type}} &
\multirow{3}{*}{\makecell{Evaluator}} &
\multicolumn{5}{c}{Non-Dialogue} & \multicolumn{4}{c}{Dialogue} \\ 
\cmidrule(lr){3-7} \cmidrule(lr){8-11}
& & 
News & 
Lifestyle & 
Report & 
\makecell{Med lit} & 
Sci-fi & 
\makecell{Daily Life} & 
Booking & 
Interview & 
Meeting \\ 
\midrule
\multirow{2}{*}{QA-based} &
  UniEval\textsubscript{faith} &
  0.11 &
  ~~0.54* &
  -0.22* &
  \!\!-0.09 &
  -0.26* &
  0.12 &
  - &
  ~~0.17* &
  0.06 \\
 &
  QAFactEval &
  ~~0.14* &
  ~~0.45* &
  \!\!-0.07 &
  ~~0.19* &
  -0.15* &
  ~~0.22* &
  0.06 &
  \!\!-0.04 &
  0.03 \\
\midrule
\multirow{3}{*}{NLI-based} &
SummaC\textsubscript{Conv} &
  0.07 &
  ~~0.13* &
  -0.15* &
  \!\!-0.13 &
  \!\!-0.08 &
  \!\!-0.11 &
  \!\!-0.03 &
  -0.21* &
  \!\!-0.08 \\
 &
  AlignScore &
  ~~0.18* &
  ~~0.32* &
  ~~0.17* &
  0.09 &
  ~~0.26* &
  ~~0.33* &
  0.12 &
  0.09 &
  ~~0.38* \\
 &
  MiniCheck &
  ~~0.24* &
  ~~0.69* &
  \!\!-0.02 &
  0.13 &
  \!\!-0.10 &
  ~~0.22* &
  ~~0.30* &
  ~~0.15* &
  0.05 \\
\midrule
\multirow{3}{*}{LLM-based} &
  G-Eval\textsubscript{faith} &
  ~~\textbf{0.65*} &
  ~~\textbf{0.72*} &
  ~~0.41* &
  \!\!-0.14 &
  ~~\textbf{0.41*} &
  ~~\textbf{0.68*} &
  ~~\textbf{0.62*} &
  ~~0.48* &
  ~~\textbf{0.60*} \\
 &
  G-Eval+\textsubscript{faith} &
  ~~0.63* &
  ~~0.57* &
  ~~\textbf{0.46*} &
  ~~\textbf{0.55*} &
  ~~0.38* &
  ~~0.46* &
  ~~0.59* &
  ~~\textbf{0.52*} &
  ~~0.53* \\          
&
  FactScore\textsubscript{faith} &
  0.07 &
  \!\!-0.02 &
  \!\!-0.11 &
  0.10 &
  0.00 &
  \!\!-0.06 &
  \!\!-0.13 &
  0.10 &
  0.05\\     
\bottomrule
\end{tabular}
\vspace*{-0.25cm}
\caption{
Agreement with human scores in \textbf{faithfulness} evaluation across nine domains (*: p-value < 0.05). For LLM-based methods, G-Eval, G-Eval+, and FactScore are the summary, sentence, and atomic level evaluators.
}
\label{tab:faith_evaluator_score_table_summary_level}
\end{center}
\vspace*{-0.3cm}
\end{table*}

\begin{table*}[t]
\begin{center}
\scriptsize
\setlength{\tabcolsep}{8pt}
\renewcommand{\arraystretch}{1.1}
\begin{tabular}{ccccccccccc}
\toprule
\multirow{3}{*}{\makecell{Model \\ Type}} &
\multirow{3}{*}{\makecell{Evaluator}} &
\multicolumn{5}{c}{Non-Dialogue} & \multicolumn{4}{c}{Dialogue} \\ 
\cmidrule(lr){3-7} \cmidrule(lr){8-11}
& & 
News & 
Lifestyle & 
Report & 
\makecell{Med lit} & 
Sci-fi & 
\makecell{Daily Life} & 
Booking & 
Interview & 
Meeting \\ 
\midrule
QA-based                        
& UniEval\textsubscript{coh}& ~~0.18* & 0.04    & 0.06      & ~~0.17*  & ~~0.15*    & 0.05      & -        & ~~0.19*    & ~~0.24*       \\
\midrule
\multirow{2}{*}{NLI-based} 
& Lite\textsuperscript{3}Pyramid        & ~~0.35* & \!\!-0.11   & ~~0.36*     & ~~0.38*  & ~~0.57*    & ~~0.36*     & -        & ~~0.25*    & ~~0.14*       \\
& A\textsuperscript{3}CU
& ~~0.43* & 0.06    & ~~0.29*     & 0.13   & ~~0.42*    & ~~0.31*     & -        & ~~0.24*    & 0.08        \\
\midrule
\multirow{2}{*}{LLM-based}      
  & G-Eval\textsubscript{coh}           
  & ~~0.35* & ~~0.45*   & ~~0.47*     & ~~0.57*  & ~~0.56*    & ~~0.41*     & ~~0.31*  
  & ~~0.55*    & ~~0.63*       \\
  
  & G-Eval+\textsubscript{com}           & ~~\textbf{0.57*} & ~~\textbf{0.61*}   & ~~\textbf{0.59*}     & ~~\textbf{0.65*}  & ~~\textbf{0.68*}    & ~~\textbf{0.56*}     & ~~\textbf{0.32*}    & ~~\textbf{0.63*}    & ~~\textbf{0.66*}  \\     
\bottomrule
\end{tabular}
\vspace*{-0.25cm}
\caption{Agreement with human scores in  \textbf{completeness} evaluation across nine domains (*: p-value < 0.05). For UniEval and G-Eval, we use their coherence scores since completeness dimension is not directly supported.
}
\label{tab:comp_evaluator_score_table_summary_level}
\end{center}
\vspace*{-0.3cm}
\end{table*}

\begin{table*}[t!]
\begin{center}
\scriptsize
\setlength{\tabcolsep}{8.5pt}
\renewcommand{\arraystretch}{1.1}
\begin{tabular}{ccccccccccc}
\toprule
\multirow{3}{*}{\makecell{Model \\ Type}} &
\multirow{3}{*}{\makecell{Evaluator}} &
\multicolumn{5}{c}{Non-Dialogue} & \multicolumn{4}{c}{Dialogue} \\ 
\cmidrule(lr){3-7} \cmidrule(lr){8-11}
& & 
News & 
Lifestyle & 
Report & 
\makecell{Med lit} & 
Sci-fi & 
\makecell{Daily Life} & 
Booking & 
Interview & 
Meeting \\ 
\midrule
QA-based &
  UniEval\textsubscript{rel} &
  0.06 &
  0.05 &
  0.02 &
  \!\!-0.01 &
  ~~0.30* &
  0.08 &
  - &
  ~~0.23* &
  ~~0.24* \\
\midrule
\multirow{2}{*}{LLM-based} &
  G-Eval\textsubscript{rel} &
  0.02 &
  ~~\textbf{0.39*} &
  ~~\textbf{0.18*} &
  0.09 &
  ~~\textbf{0.51*} &
  ~~0.33* &
  ~~\textbf{0.2*} &
  ~~0.46* &
  ~~0.45* \\
 &
  G-Eval+\textsubscript{con}&
  0.11 &
  ~~\textbf{0.39*} &
  ~~0.17* &
  ~~\textbf{0.24*} &
  ~~0.44* &
  ~~\textbf{0.36*} &
  0.02 &
  ~~\textbf{0.49*} &
  ~~\textbf{0.45*} \\
\bottomrule
\end{tabular}
\vspace*{-0.25cm}
\caption{Agreement with human scores in \textbf{conciseness} evaluation across nine domains (*: p-value < 0.05).  For UniEval and G-Eval, we use their relevance scores since conciseness dimension is not directly supported.
}
\label{tab:conc_evaluator_score_table_summary_level}
\end{center}
\vspace*{-0.6cm}
\end{table*}

Appendix~\ref{app:hallu_full_result} provides the detailed results without the aggregation for the faithfulness, completeness, conciseness, and composite scores.

\subsection{Impact of PII Redaction.}

We investigate how PII redaction impacts summarization quality using recent summarizers. This is a crucial aspect, because it is very common in industrial use cases, such as call centers and legal service. To construct a redacted dataset, we select a subset of \algname{} -- 25 hallucinogenic texts from the MultiWOZ (booking) data which contain significant amounts of PII-related entities, including phone numbers and addresses. We manually redact the entities by replacing them with their corresponding category name, \emph{i.e.}, \textsc{<phone-number-1>}. On average, eight entities are redacted per input dialogue. See Appendix~\ref{app:redaction_protocol} for the detailed protocol.

Table \ref{tab:redaction_results_abc} shows the faithfulness scores before and after PII redaction into input texts, where the top-2 summarizers are selected from each category. PII redaction negatively affects the faithfulness of the all summarizers. \textbf{The non-LLMs are more susceptible to PII redaction than the open-source and proprietary LLMs.} This drop is attributed to filling in the masked entity with an entity either not present in the input text or incorrectly presented.

\subsection{Factuality Error Analysis}

\begin{figure}[t!]
\begin{center}
\vspace*{-0.25cm}
\includegraphics[width=8cm]{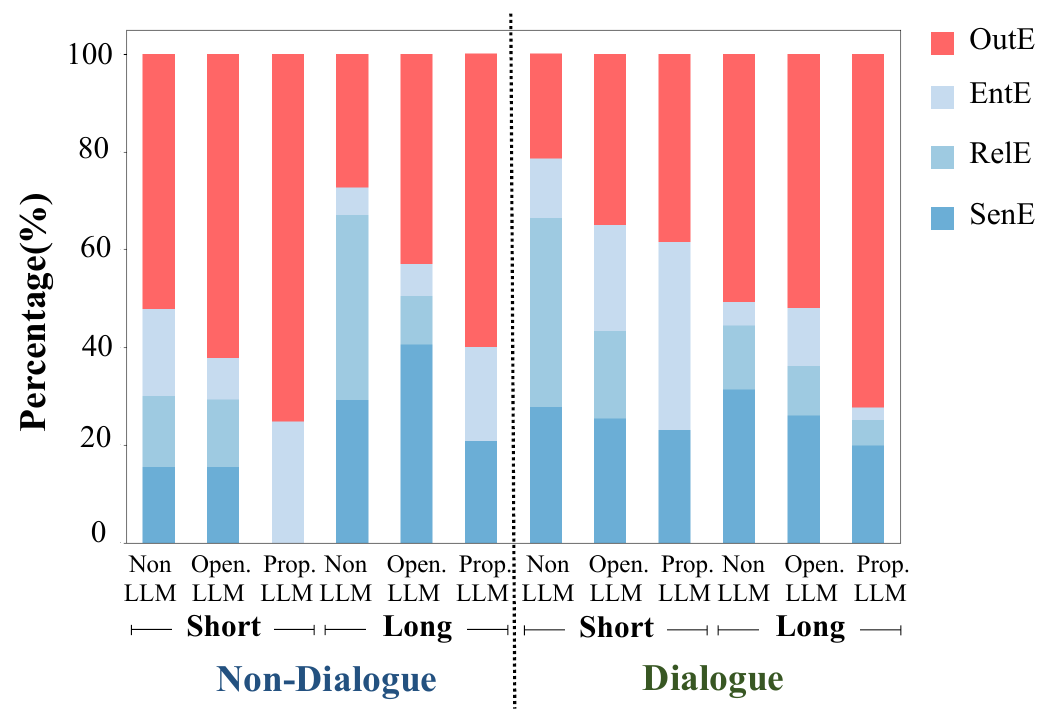}
\end{center}
\vspace*{-0.45cm}
\caption{Error distribution by varying input contexts for each summarizer category, showing OutE (out-of-article error), EntE (entity-error), RelE (relation-error), and SenE (sentence-error). \textcolor{red}{Red} color indicates extrinsic errors, while \textcolor{blue}{blue} tones denotes intrinsic errors.}
\label{fig:error_type_group}
\vspace*{-0.45cm}
\end{figure}

In Figure \ref{fig:error_type_group}, we examine how the distribution of error types varies across different input contexts for each summarizer group. 
\textbf{The proprietary LLMs exhibit a lower rate of intrinsic errors, while the non-LLMs exhibit a higher rate of them in all input contexts.}
Notably, the proprietary LLMs show no relation errors across most context types except in long dialogue texts. 
This suggests that the higher faithfulness scores of the proprietary LLMs in Table \ref{tab:overall_score} are likely due to their much lower intrinsic error rates compared with the others.

\section{Benchmarking Auto-Evaluators }\label{sec:exp-evaluator}

\noindent\textbf{Evaluator Selection.}
We benchmark SOTA automated evaluators on \algname{}. The set of compared evaluators varies according to the target evaluation dimension. For faithfulness, we include \emph{QA-based} models: UniEval \cite{zhong2022towards} and QAFactEval \cite{fabbri2022qafacteval}, \emph{NLI-based} models: Summac-Conv \cite{laban2022summac}, AlignScore \cite{zha2023alignscore}, and MiniCheck \cite{tang2024minicheck}, and \emph{LLM-based} models at various levels of granularity: G-Eval \cite{liu2023geval} for summary level, G-Eval{+}\footnote{We adjusted G-Eval's prompts to align with the  granularities and dimensions of \algname{}, renaming it G-Eval{+}. See Appendix~\ref{app:autoeval_details} for the tuned prompts.} for sentence level and FactScore \cite{min2023factscore} for atomic level. 
For completeness, we include NLI-based models: Lite\textsuperscript{3}Pyramid \cite{zhang2021finding}, A\textsuperscript{3}CU \cite{liu2023towards}, along with UniEval and G-Eval+.
For conciseness, we include the models supporting the evaluation of relevance, such as UniEval and G-Eval, and our G-Eval+ tailored for conciseness. We use GPT-4-turbo for all LLM-based models.

\smallskip\smallskip
\noindent\textbf{Evaluation Metric.}  Following prior works \cite{liu2023geval, fu2023gptscore}, we compare the estimated scores with the ground-truth human scores in \algname{}. We report their Pearson correlation based on the \emph{summary-level} percentage score of each evaluation dimension. See Appendix \ref{app:eval_calculation} for the details of measurements and 
Appendix \ref{app:benchmark_system} more results using system-level measurement.

\subsection{Alignment with Human Score}

Tables \ref{tab:faith_evaluator_score_table_summary_level}--\ref{tab:conc_evaluator_score_table_summary_level} report the agreement between automated evaluators and humans in faithfulness, completeness, and conciseness. 
In the booking domain, evaluators that require reference summaries do not report results since these summaries are unavailable.
In general, the LLM-based evaluators, G-Eval+, show the highest agreement in all dimensions. The agreement with the human scores appears to vary across different data domains. 

In the faithfulness evaluation (Table \ref{tab:faith_evaluator_score_table_summary_level}), \textbf{the increasing granularity over the three LLM-based methods do not guarantee improved performance.} Specifically, the atomic level evaluator, FactScore, does not perform well due to the difficulty in atomic fact generation -- often producing numerous overlapping or factually incorrect atomic facts, which can cascade into further errors by LLMs.
Moreover, contrary to prior findings that NLI-based evaluators perform well\,\cite{tang2024tofueval}, they show much lower agreement compared to G-Eval when evaluating faithfulness on our hallucinogenic texts. Hence, \textbf{non-LLM evaluators tend to perform poorly at verifying LLM-generated hallucinations.}

{Additionally, \textbf{the agreement with human faithfulness evaluation is significantly affected by domain characteristics.} In particular, QA- and NLI-based methods, which require training on specific data, lack domain generalization in automatic evaluation. They exhibit negative correlations with human judgements on more than half of the domains in faithfulness evaluation. In contrast, the LLM-based evaluators (G-Eval and G-Eval+) show a significantly higher positive correlation compared to the other evaluators.}

In the completeness and conciseness evaluation (Tables \ref{tab:comp_evaluator_score_table_summary_level}--\ref{tab:conc_evaluator_score_table_summary_level}), for the LLM-based methods, \textbf{employing prompts fine-tuned for target dimensions leads to higher agreements in general}, \emph{i.e.}, G-Eval $\rightarrow$ G-Eval+.
Lastly, we observe a considerable performance discrepancy of SOTA automated evaluators in the conciseness dimension compared to the faithfulness and completeness dimensions. While the LLM-based evaluators show a fairly high agreement of up to 0.68 in evaluating faithfulness and completeness, they exhibit considerably lower agreement of up to 0.51 for conciseness evaluation. 
This highlights that \textbf{evaluating conciseness could be harder than other dimensions} in the text summarization task.

{In general, LLM-based evaluators (except for FactScore in faithfulness) achieve higher agreement with human judgments compared to QA- and NLI-based evaluators across all three dimensions. However, they still fall short in certain domains as automated evaluators. If a threshold of 0.50 is set for satisfactory correlation with human judgments, G-Eval+ fails to meet this standard in the Report and Sci-fi domains for faithfulness evaluation, the Booking domain for completeness evaluation, and most domains for conciseness evaluation.}

\vspace*{-0.1cm}
\section{Conclusion}\label{sec:con}
\vspace*{-0.2cm}

{We introduce \algname{}, a benchmark dataset featuring hallucinogenic texts from nine domains, spanning non-dialogue to dialogue and short to long texts, paired with their summaries generated by nine recent summarizers. Built using a data creation pipeline with AI assistance, \algname{} includes high-quality, fine-grained human annotations that enable in-depth studies on the multi-dimensional performance of summarizers.}
{Additionally, based on our benchmark, we provide a thorough assessment of automated evaluators for text summarization, revealing weaknesses related to specific domains and evaluation dimensions.}

\vspace*{-0.1cm}
\section*{Limitations}\label{sec:lim}
\vspace*{-0.2cm}

Our work has some limitations. First, although \algname{} covers three comprehensive evaluation dimensions for summarization quality, additional dimensions like harmfulness or bias could enhance the nuanced assessment of summaries. Second, the generation of key-facts could be refined by developing tailored strategies for different domains to better extract domain-specific key-facts. Third, since the importance of key-facts can vary, key-fact alignment could measure the completeness and conciseness of summaries more precisely by considering the relative importance of each key-fact. Finally, although we achieve high IAA for human annotations, the IAA for long texts remains lower than for short texts. Future research is required to refine annotation strategies for long texts. Despite these challenges, we hope our work will give valuable insights into the field of text summarization and foster the development of a more advanced automated evaluators.
\vspace*{-0.1cm}
\section*{Ethics Statement}\label{sec:ethics}
\vspace*{-0.2cm}

We actively addressed annotators' queries during the annotation process, ensuring faithful communication. Annotators were compensated at a rate 50\% above the average American minimum wage and received bonuses for consistent, high-quality work. Our dataset excludes any information that could potentially disclose the annotators' personal details.

\vspace*{-0.1cm}
\section*{Scientific Artifacts}\label{sec:sci}
\vspace*{-0.2cm}

We utilized nine language models to generate summaries on \algname{}. Apart from the paid APIs like OpenAI and AWS Bedrock, we used readily available checkpoints on Huggingface. All the details are summarized in Table \ref{tab:model_checkpoint}.
\vspace*{-0.1cm}
\section*{Acknowledgements}\label{sec:ack}
\vspace*{-0.2cm}

This work was supported by Institute of Information \& communications Technology Planning \& Evaluation (IITP) grant funded by the Korea goverment (MSIT) (No. RS-2024-00445087, Enhancing AI Model Reliability Through Domain-Specific Automated Value Alignment Assessment). Additionally, this work was partly supported by the National Research Foundation of Korea (NRF) grant funded by the Korea government (MSIT) (No. RS-2024-00334343) and the National Research Foundation of Korea (NRF) funded by  Ministry of Science and ICT (NRF-2022M3J6A1063021).


\appendix
\clearpage

\section{Summary of the Source datasets}\label{app:datasets}

\begin{table*}
\begin{center}
\scriptsize
\renewcommand{\arraystretch}{1.1}
\setlength{\tabcolsep}{11pt}
\begin{tabular}{c|c|c|c|c|c|c}
\toprule
Dataset &
Type &
\makecell{Text \\ Length} &
Domain &
\makecell{Text \\ Word count \\ (Min -- Max)} &
\makecell{Summary \\ Word count \\ (Min -- Max)} &
\makecell{Key-fact Count \\ (Min -- Max)} \\ \midrule
CNNDM       & \multirow{5}{*}{\makecell{\\ \\ Non-Dialogue}} & \multirow{2}{*}{\makecell{Short}} & News         & \makecell{674 (227 -- 1,231)} & \makecell{105.7 (19 -- 420)} & \makecell{11.4 (5 -- 17)} \\ 
\cmidrule(lr){1-1} \cmidrule(lr){4-7}
WikiHow     &                                        &                        & Lifestyle    & \makecell{65.6 (21 -- 151)}  & \makecell{43.6 (5 -- 152)}   & \makecell{4.2 (1 -- 13)}   \\ 
\cmidrule(lr){1-1} \cmidrule(lr){3-7}
GovReport   &                                        & \multirow{3}{*}{\makecell{\\ Long}}  & Report       & \makecell{6,263.7 (2,429 -- 10,462)} & \makecell{155.9 (4 -- 764)} & \makecell{17.8 (11 -- 20)} \\ 
\cmidrule(lr){1-1} \cmidrule(lr){4-7}
PubMed      &                                        &                        & Medical      & \makecell{3,220.4 (1,204 -- 5,586)} & \makecell{165.1 (4 -- 2,349)} & \makecell{17.7 (11 -- 20)} \\ 
\cmidrule(lr){1-1} \cmidrule(lr){4-7}
SQuALITY    &                                        &                        & Sci-fiction  & \makecell{6,084.8 (4,782 -- 6,720)} & \makecell{110.1 (2 -- 312)} & \makecell{12.4 (7 -- 18)} \\ 
\midrule
DialogSum   & \multirow{4}{*}{\makecell{\\ \\ Dialogue}}              & \multirow{2}{*}{\makecell{ Short}} & Daily Life   & \makecell{154.2 (63 -- 287)} & \makecell{43.4 (7 -- 128)} & \makecell{6.8 (3 -- 14)} \\ 
\cmidrule(lr){1-1} \cmidrule(lr){4-7}
MultiWOZ    &                                        &                        & Booking      & \makecell{252 (118 -- 349)} & \makecell{64.1 (10 -- 125)} & \makecell{8.2 (4 -- 13)} \\ 
\cmidrule(lr){1-1} \cmidrule(lr){3-7}
MediaSum    &                                        & \multirow{2}{*}{\makecell{\\ Long}}  & Interview    & \makecell{1,635.2 (631 -- 2,978)} & \makecell{113.1 (17 -- 572)} & \makecell{12.6 (6 -- 19)} \\ 
\cmidrule(lr){1-1} \cmidrule(lr){4-7}
MeetingBank &                                        &                        & Meeting      & \makecell{978.3 (293 -- 3,389)} & \makecell{89.3 (7 -- 713)} & \makecell{9.2 (5 -- 13)} \\ \midrule
\multicolumn{4}{c|}{UniSummEval} & \makecell{2,092 (21-10,462)} & \makecell{133 (2 --  2,349)} & \makecell{11.1 (1-20)} \\ \bottomrule
\end{tabular}
\caption{Summary of the nine datasets in \algname{}: the average word count of texts and summaries, and the number of key facts, with their respective minimum and maximum ranges in parentheses. \algname{} sampled 25 hallucinogenic texts from each dataset.}
\label{table:source datasets}
\end{center}
\vspace{-0.3cm}
\end{table*}

Table \ref{table:source datasets} provides detailed statistics on the nine source datasets and their generated summaries/key-facts, where datasets with an average word count of more than 900 are classified as long texts. The selection of the source datasets is to cover various domains, with a balanced distribution of text types (dialogue, non-dialogue) and lengths (long, short). {Our benchmark contains a total of 225 source documents, with each domain equally containing 25 documents.}






\section{Model Settings and Prompts}

\subsection{Summary Generation Details}\label{app:summary_generation_and_prompts}

\begin{figure}[ht!]
\begin{center}
\vspace*{-0.1cm}
\includegraphics[width=7.7cm]{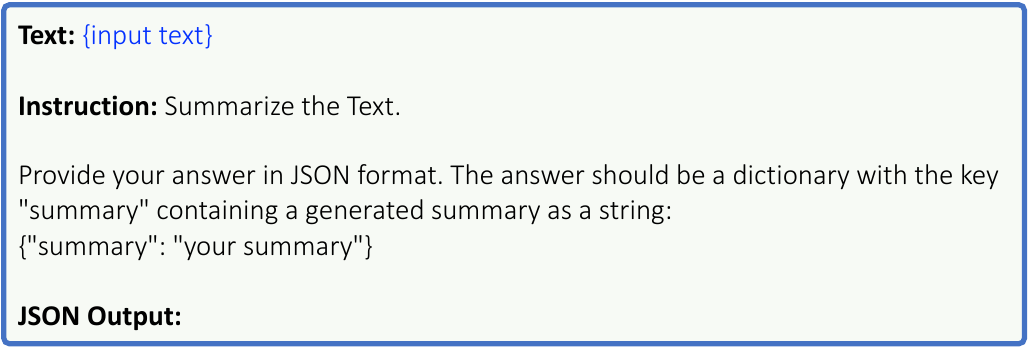} 
\vspace*{-0.5cm} 
\end{center}
\vspace*{-0.4cm}
\caption{The prompt to generate a summary.}
\label{fig:summarization prompt}
\vspace*{-0.1cm}
\end{figure}

\begin{table}[ht!]
\footnotesize
\begin{center}
\setlength{\tabcolsep}{4.5pt}
\renewcommand{\arraystretch}{1.1}
\begin{tabular}{lll}
\toprule
\multicolumn{1}{l}{Model   Name}        & HuggingFace Checkpoints                          &  \\ \midrule
\multirow{2}{*}{BART\textsubscript{large}}    
& facebook/bart-large-cnn              &  \\
& linydub/bart-large-samsum            &  \\
\multirow{2}{*}{Flan-T5\textsubscript{large}} 
& spacemanidol/flan-t5-large-cnndm     &  \\
& oguuzhansahin/flan-t5-large-samsum   &  \\
Phi-2                          
& microsoft/phi-2                      &  \\
Mistral\textsubscript{7B-Inst}                
& mistralai/Mistral-7B-Instruct-v0.2   &  \\
Llama2\textsubscript{13B-chat}                
& meta-llama/Llama-2-13b-chat-hf        &  \\
Mixtral\textsubscript{8x7B-Inst}             
& mistralai/Mixtral-8x7B-Instruct-v0.1 &  \\
GPT-3.5\textsubscript{turbo}                  
& gpt-3.5-turbo-0125*                   &  \\
GPT-4\textsubscript{turbo}                          
& gpt-4-0125-preview*                   &  \\
Claude2.1                      
& claude-2.1*                           &  \\ 
\bottomrule
\end{tabular}
\vspace*{-0.25cm}
\caption{The checkpoints of the summarization models. *Their official APIs are used.}
\label{tab:model_checkpoint}
\end{center}
\vspace*{-0.4cm}
\end{table}

We briefly describe the settings of the summarization models in our benchmark. For the two non-LLMs, BART-large and Flan-T5-large, we choose the pre-trained models in HuggingFace model hub according to whether the domain is for dialogue or non-dialogue. We use instruction-tuned model checkpoints for the open-source LLMs and the official APIs for the proprietary LLMs. The checkpoints used for each model can be found in Table \ref{tab:model_checkpoint}.
We set the temperature to 1 and use the prompt shown in Figure \ref{fig:summarization prompt} for generating summaries across the summarizers.

\subsection{AI-assistant Details}\label{app:ai_assistant_prompt}
\begin{figure}[ht!]
\begin{center}
\vspace*{-0.1cm}
\includegraphics[width=7.7cm]{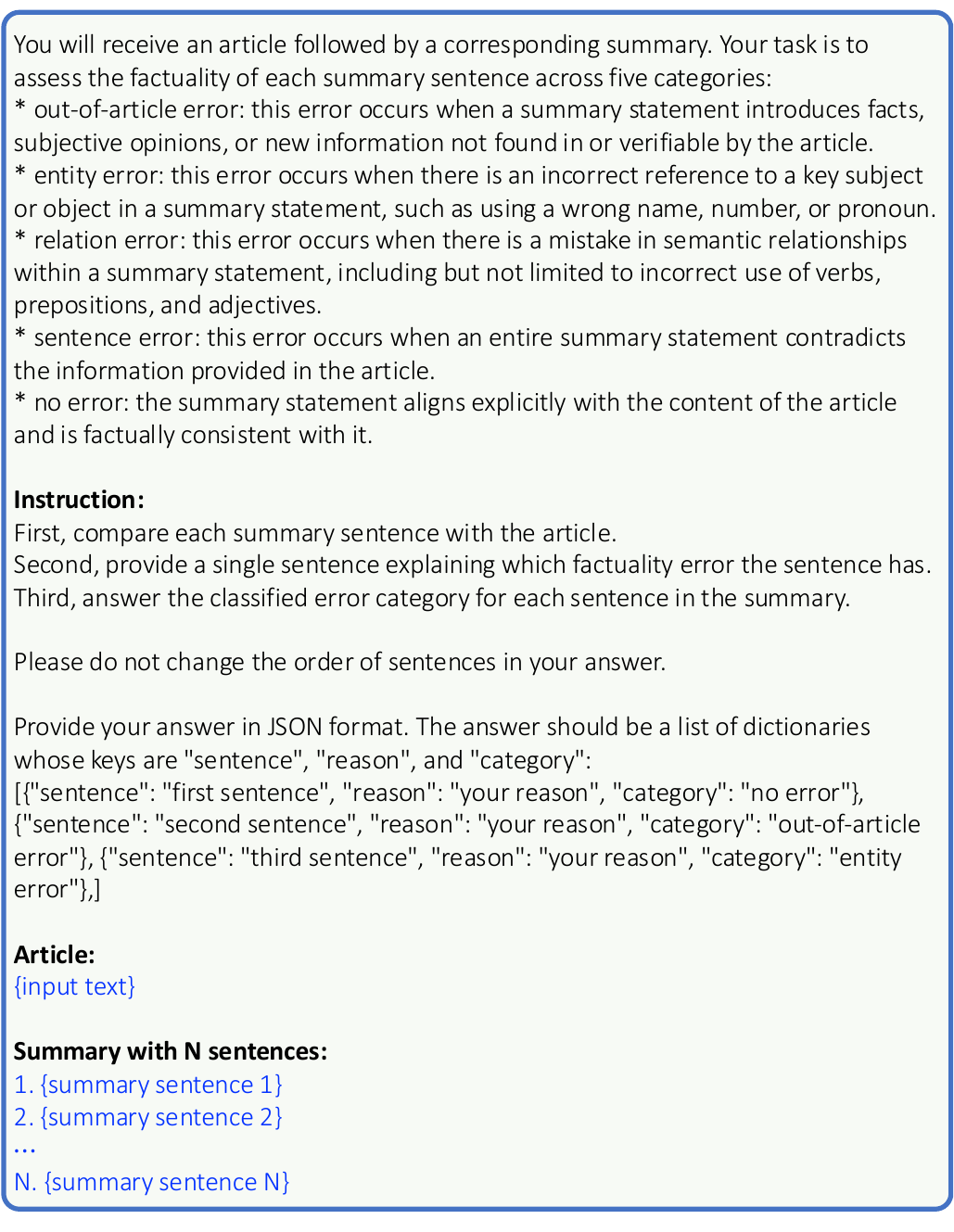} 
\vspace*{-0.5cm} 
\end{center}
\vspace*{-0.5cm}
\caption{The prompt to generate AI evaluations on faithfulness.}
\label{fig:ai_assist_prompt}
\vspace*{-0.1cm}
\end{figure}

We conduct LLM-based summary faithfulness evaluation to (1) select hallucinogenic texts and (2) provide machine-based reasoning to aid in the human annotation of fact verification. {We modify the factual error types originally used in the prompt of FineSurE\,\cite{song2024finesure}, reducing them from nine to five. This makes the annotation task more intuitive and feasible for human annotators.}
The prompt to generate AI faithfulness evaluations is provided in Figure \ref{fig:ai_assist_prompt}.
{\paragraph{Reliability of AI Evaluation} We use Claude 2.1 to generate AI faithfulness evaluations. To ensure the reliability of these evaluations, we conduct an automated faithfulness evaluation using an additional SOTA LLM, Llama3-70B-Instruct, on the entire pool of hallucinogenic texts. As a result, we find that 98.4\% (1,180 out of 1,199) of the texts are confirmed as hallucinogenic by both models, suggesting that the bias may be insignificant.}



\subsection{Key-fact Generation Details}\label{app:key-fact_extraction_prompt}

\begin{figure}[ht!]
\begin{center}
\vspace*{-0.25cm}
\includegraphics[width=7.7cm]{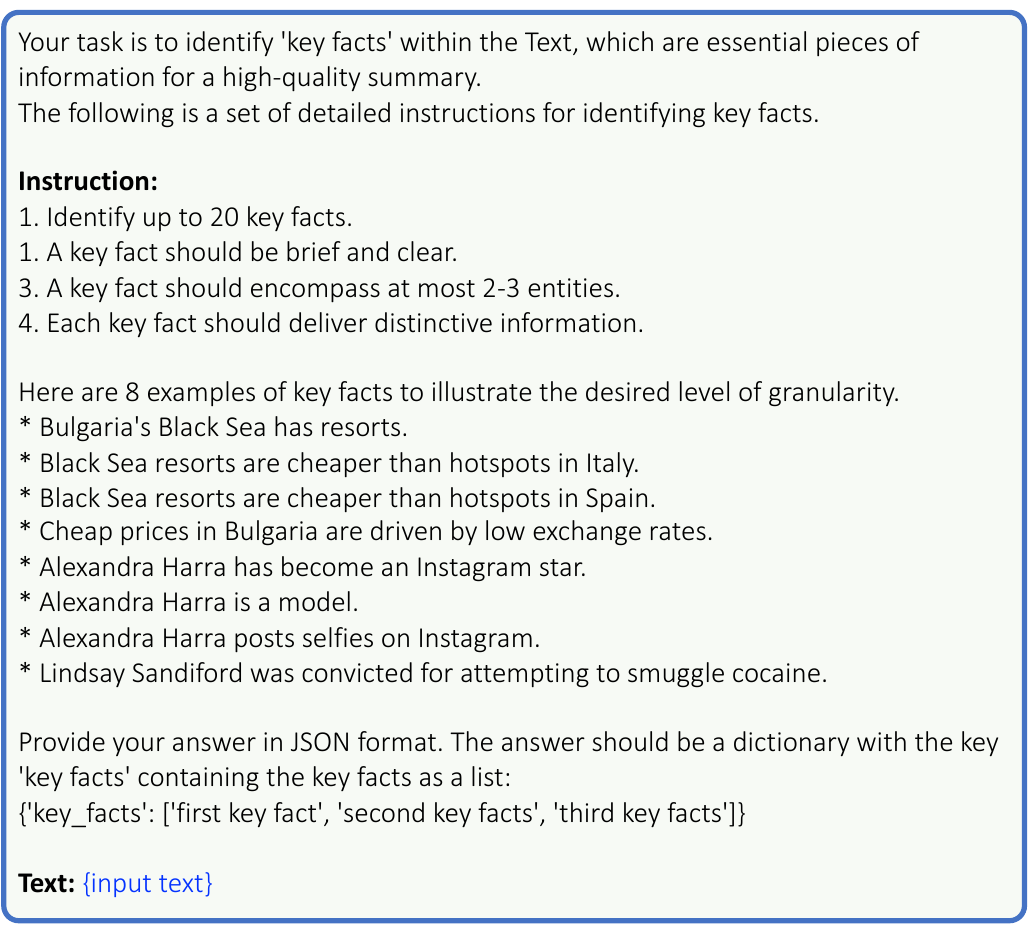} 
\vspace*{-0.5cm} 
\end{center}
\vspace*{-0.45cm}
\caption{The prompt to generate key-facts.}
\label{fig:keyfact_extraction_prompt}
\vspace*{-0.1cm}
\end{figure}
\begin{figure}[ht!]
\begin{center}
\vspace*{-0.15cm}
\includegraphics[width=7.7cm]{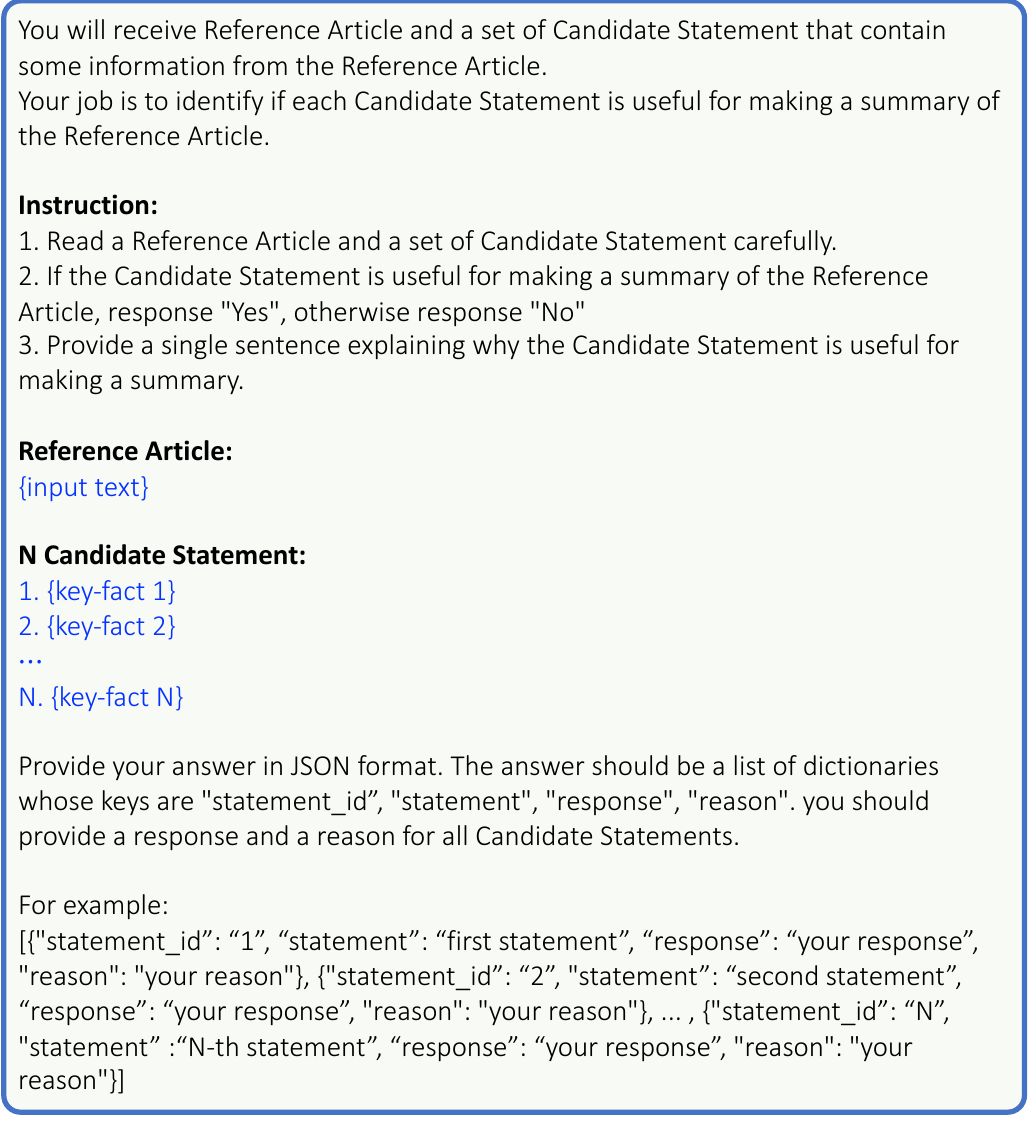} 
\vspace*{-0.5cm} 
\end{center}
\vspace*{-0.45cm}
\caption{The prompt to cross-validate key-facts.}
\label{fig:keyfact_removal_prompt}
\vspace*{-0.2cm}
\end{figure}

We extract an initial set of key-facts from each source text using GPT-4-turbo with the prompt described in  Figure \ref{fig:keyfact_extraction_prompt}. To ensure the quality of the initial key-facts, we cross-validate them using GPT-3.5-turbo, Claude 2.1, and Mixtral-8x7B with the prompt described in Figure \ref{fig:keyfact_removal_prompt}.

\subsection{Auto-Evaluators Details}\label{app:autoeval_details}

For the non-LLM evaluators such as QA- and NLI-based ones, we choose their default models or load model checkpoints that demonstrated the best performance in the paper. All other settings, including hyperparameters and prompts, are kept as provided in the original papers. In the case of FactScore, reference retrieval from the knowledge base is unnecessary; the evaluator assesses the faithfulness of the summary based on the input text alone.

\begin{figure}[t!]
\begin{center}
\vspace*{-0.1cm}
\includegraphics[width=7.7cm]{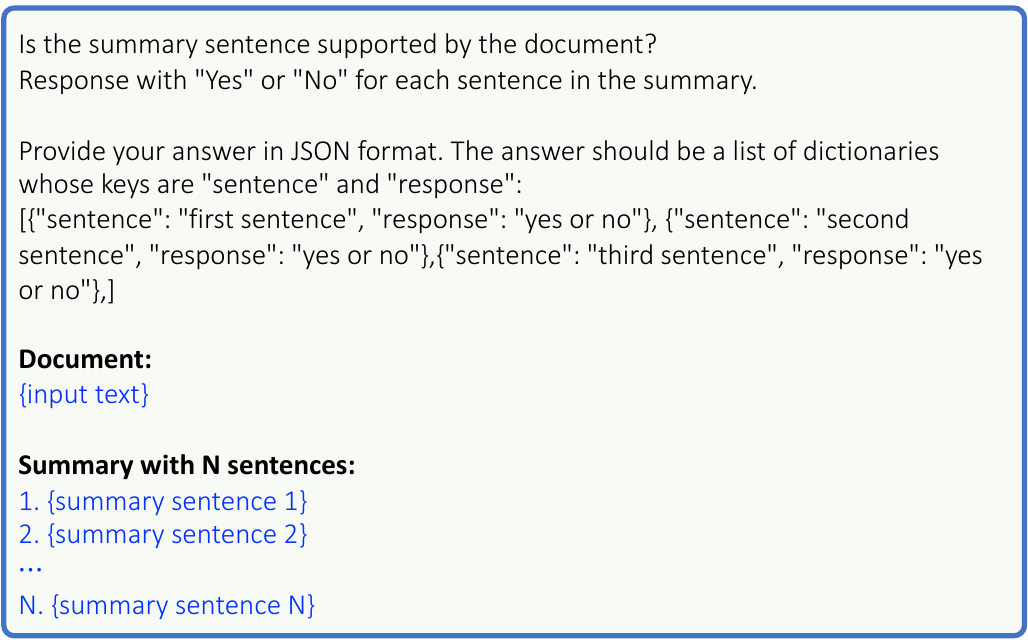} \vspace*{-0.5cm} 
\end{center}
\vspace*{-0.25cm}
\caption{G-Eval+ prompt tailored for sentence level in faithfulness evaluation.}
\label{fig:g_eval_faithfulness_sentence_prompt}
\vspace*{-0.1cm}
\end{figure}

\begin{figure}[ht!]
\begin{center}
\vspace*{-0.25cm}
\includegraphics[width=7.7cm]{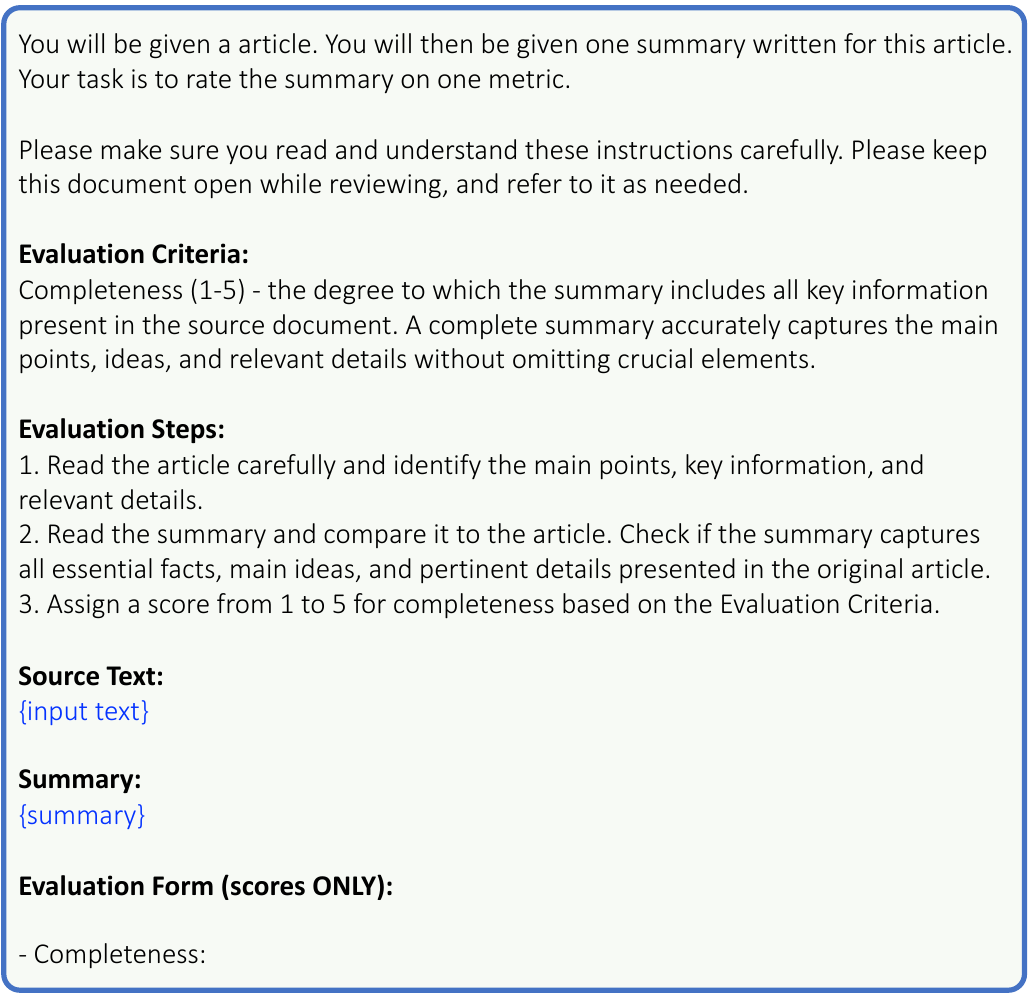} \vspace*{-0.5cm} 
\end{center}
\vspace*{-0.4cm}
\caption{G-Eval+ prompt tailored for completeness.}
\label{fig:g_eval_completeness_prompt}
\vspace*{-0.2cm}
\end{figure}

\begin{figure}[ht!]
\begin{center}
\vspace*{-0.25cm}
\includegraphics[width=7.7cm]{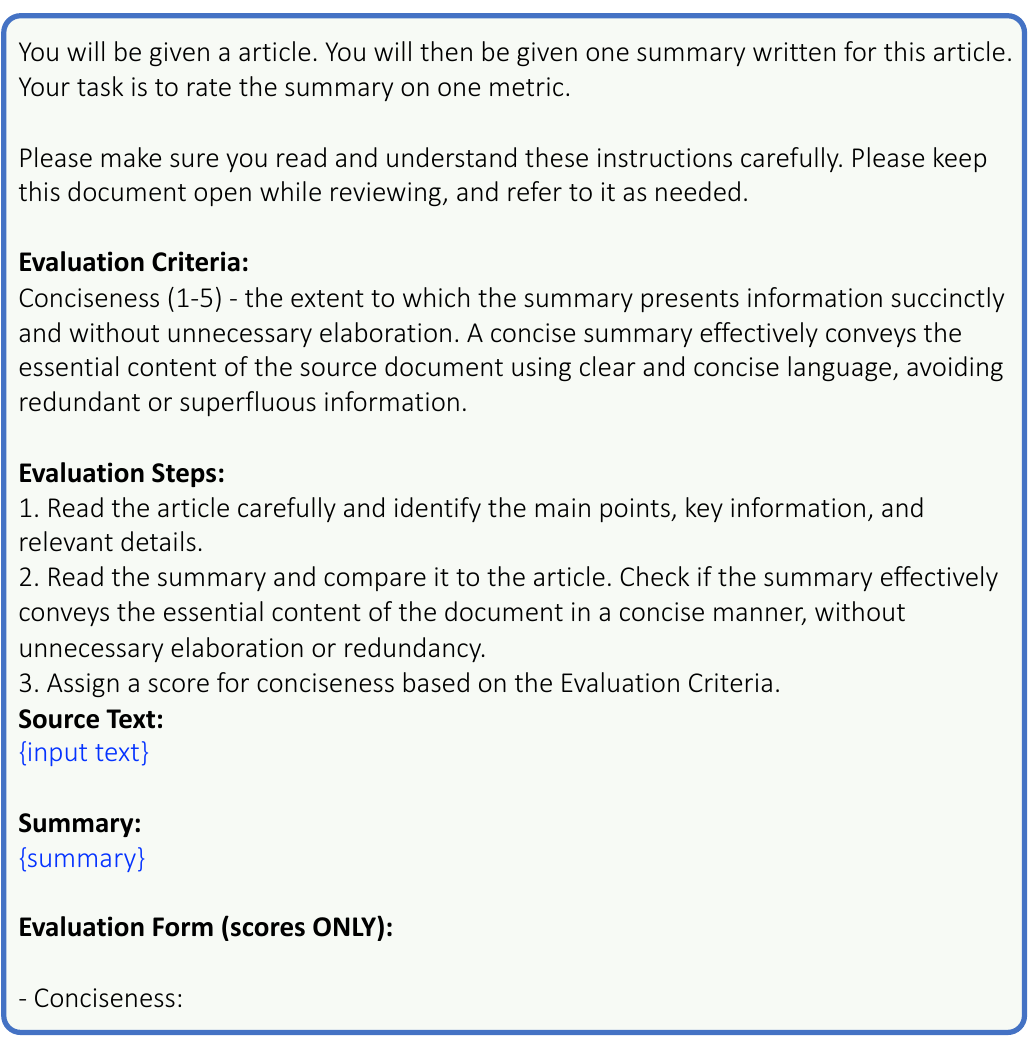} \vspace*{-0.5cm} 
\end{center}
\vspace*{-0.4cm}
\caption{G-Eval+ prompt tailored for conciseness.}
\label{fig:g_eval_conciseness_prompt}
\vspace*{-0.6cm}
\end{figure}

We also use customized G-Eval prompts, referred to as G-Eval+, tailored to our three evaluation dimensions: faithfulness, completeness, and conciseness. The original G-Eval prompt alone does not perfectly align with these dimensions or their granularities. Therefore, we develop specific prompts for each dimension. For faithfulness, we evaluate at the sentence level, while for completeness and conciseness, we use prompts aligned with the evaluator's criteria. The detailed prompts for faithfulness, completeness, and conciseness are shown in Figures \ref{fig:g_eval_faithfulness_sentence_prompt} --\ref{fig:g_eval_conciseness_prompt}, respectively.


\section{Factual Error Types}\label{app:factual_error_types}

Table \ref{tab:errortype_example} presents detailed descriptions and examples of the factual error types used for the fact verification annotation. The taxonomy is based on a modified version of the taxonomy suggested by \citet{mishra2024finegrained}.

\begin{table*}[t]
\scriptsize
\centering
\begin{tabular}{cll}
\hline
\multicolumn{3}{c}{Example Source Text} \\ \hline
\multicolumn{3}{l}{\begin{tabular}[c]{@{}l@{}}In the heart of the bustling city, nestled inside a park, stands the historic Jefferson Library. Built in 1910, this architectural marvel houses a vast collection \\ of rare books, manuscripts, and artifacts, attracting scholars and history enthusiasts from around the world. Its grand facade and ornate interiors make it a \\ beloved landmark, reflecting the city's rich cultural heritage and commitment to education.\end{tabular}} \\ \hline
Error Type & \multicolumn{1}{c}{Description} & \multicolumn{1}{c}{Example Summary Sentence} \\ \hline
Out-of-Article Error & \begin{tabular}[c]{@{}l@{}}This error occurs when a summary sentence introduces facts, subjective opinions, \\ or biases that cannot be verified or confirmed by the source text.\end{tabular} & \begin{tabular}[c]{@{}l@{}}The Jefferson Library was \textbf{the first library} \\ \textbf{to offer online book lending services.}\end{tabular} \\
Entity Error & \begin{tabular}[c]{@{}l@{}}This error involves incorrect or misrepresented entities (such as names, numbers,\\  or main subjects) within the summary sentence.\end{tabular} & \begin{tabular}[c]{@{}l@{}}The Jefferson \textbf{School} houses a vast collection \\ of rare books.\end{tabular} \\
Relation Error & \begin{tabular}[c]{@{}l@{}}This error arises from incorrect semantic relationships within a summary sentence, \\ such as wrong verbs, prepositions, or adjectives, which misrepresent the \\ relationship between entities.\end{tabular} & The Jefferson Library is located \textbf{beside} a park. \\
Sentence Error & \begin{tabular}[c]{@{}l@{}}This error occurs when a summary sentence entirely contradicts the information \\ in the source text, requiring significant revision or removal.\end{tabular} & \begin{tabular}[c]{@{}l@{}}The Jefferson Library is \textbf{a modern structure} \\ \textbf{with minimalist architecture.}\end{tabular} \\ \hline
\end{tabular}
\vspace*{-0.2cm}
\caption{Descriptions and examples of factual error types. The parts of each summary sentence that are relevant to the specific error type are highlighted in bold.}
\label{tab:errortype_example}
\vspace*{-0.2cm}
\end{table*}

\section{Detailed Calculation of the Benchmark Scores}
\subsection{Summarization Performance Calculation}\label{app:summary_performance_calculation}


After collecting annotations at a fine-grained level, the scores can be aggregated at a summary-level percentage score for all three dimensions, following the recent work by \citet{song2024finesure}.


For a document D, let $S = \{s_1,...,s_N\}$ be the list of generated summaries with $N$ sentences. 
Based on the result of faithfulness annotation, we can identify $S_{fact} \subseteq S$, a subset of S that are annotated as having no factual error. Consequently, the percentage score of faithfulness of summary $S$ is calculated by:
\begin{equation}
{Faithfulness}(D, S) = |S_{fact}| / |S|. 
\label{eq:faithfulness_score}
\end{equation}

Let $K=\{k_1, \dots, k_M\}$ be the set of key facts, where $M$ is the total number of key facts. Based on the result of the key fact alignment,  we can define a bipartite graph $M=(K, S, E)$, where $E$ consists of edges $\{(k, s):  k \rightarrow s~|~k \in K \wedge s \in S \}$ with $k \rightarrow s$ signifying that key fact $k$ is labelled as being present in summary sentence $s$. The completeness and conciseness score for summary $S$ are then calculated as a percentage score by:
{
\begin{equation}
\begin{gathered}
\!\!\!\!{Completeness}(K, S)\! =\! \big|\{k|(k,s)\!\in E\}\big| / |K|\!\!\\
\!\!\!\!{Conciseness}(K, S) \!= \!\big|\{s|(k,s)\!\in E\}\big| / |S|.\!\!
\end{gathered}
\label{eq:other_scores}
\end{equation}
Here, the operator $|\cdot|$ denotes the cardinality of a set. With these scores, we can quantify a summary-level score for completeness, which reflects the extent to which the key facts are incorporated into the summary. Additionally, the conciseness score measures the extent to which the summary incorporates the key facts.

\paragraph{Domain Stability Score.} 
The domain stability score quantifies how consistent a model’s performance is across the nine given domains. We first calculate the instability score by taking the difference between the maximum and minimum performance scores across these domains. The domain stability score is then determined by subtracting the instability score from a fixed upper bound of 100. The domain stability score can be calculated in terms of the four score types - faithfulness, completeness, conciseness, and the composite score.

Let \( S_i \) represent the score of the model in the \( i \)-th domain, where \( i = 1, \ldots, 9 \). The instability score $Instability$ is computed as:

\begin{equation}
Instability = \text{max}_{i} S_i - \text{min}_{i} S_i.
\end{equation}
Then, the domain stability score $DoS$ is given by:
\begin{equation}
DoS = 100 - Instability.
\end{equation}


\paragraph{Abstractiveness Score.}
Abstractiveness is quantified by calculating the ratio of novel n-grams in the summary that do not appear in the input text \cite{liu2019text, song2023enhancing}. For a summary $S$, let \( n\text{-gram}_{\text{copied}} \) be the set of n-grams that are copied from the document, and let \( n\text{-gram}_{\text{total}} \) be the set of all n-grams in the summary. Then, the ratio of novel n-grams \( N_{\text{n}} \) can be defined as:

\begin{equation}
N_{\text{n}} = 1 - \frac{n\text{-gram}_{\text{copied}}}{n\text{-gram}_{\text{total}}}.
\end{equation}

Following \citet{song2023enhancing}, the abstractiveness score for a summary $S$ is calculated as the average of the novel 1/3/5-gram ratios:

\begin{equation}
Abstractiveness(D, S) = \frac{N_{1} + N_{3} + N_{5}}{3}.
\end{equation}

\subsection{Evaluator Performance Calculation}\label{app:eval_calculation}

We calculate summary- and system-level correlation to verify the agreement between automated evaluation and human evaluation, {following the recent work\,\cite{song2024finesure, liu2023geval}.}

For summary-level evaluation, we analyze the correlations between the scores based on the human annotations and those calculated by automatic evaluators. 
Let $x_i$ be the percentage score of a evaluation based on human annotation and $y_i$ be the score generated by automated evaluators on the \(i\)-th data. Then, the summary-level correlation is calculated as follows:
\begin{equation}
{\rm Corr}\left(\left[x_1, x_2, \dots, x_n\right], \left[y_1, y_2, \dots, y_n\right]\right). 
\end{equation}
where {Corr} is a function calculating a pearson correlation coefficient.

For system-level evaluation, we aggregate the percentage scores for each summarizer across all input texts, and then calculate the rank correlation between the ranks based on human annotation results. Let \( X_{i,j} \) represent the score calculated by an automated evaluator, on an input text \( i \) for a summarizer \( j \). The aggregated score \( \overline{X}_j \) for a summarizer \( j \) is given by:

\begin{equation}
\overline{X}_j = \frac{1}{n} \sum_{i=1}^{n} X_{i,j}.
\end{equation}
Similarly, let \( Y_{i,j} \) represent 
the human score for a summarizer \( j \) for an input text \( i \). The aggregated score \( \overline{Y}_j \) for a summarizer \( j \) based on human evaluation is given by:
\begin{equation}
\overline{Y}_j = \frac{1}{n} \sum_{i=1}^{n} Y_{i,j}.
\end{equation}
Then, we make a ranking of the average percentage scores for all summarization systems. Let \( R(X_j) \) represent the rank of \( \overline{X}_j \) and \( R(Y_j) \) represent the rank of \( \overline{Y}_j \). we compute the rank correlation by:
\begin{equation}
\begin{gathered}
{\rm Spearman}\Big(\left[R(X_1), R(X_2), \dots\right], \\
\left[R(Y_1), R(Y_2), \dots\right]\Big).
\end{gathered}
\end{equation}

\section{Redaction Protocol}\label{app:redaction_protocol}

For redaction, we identify six specific categories requiring redaction: place, time, day, phone number, code, (such as reservation numbers and train numbers), and address (including postcodes). All pertinent entities within the source text are redacted using their category name and angle brackets "<>".
Furthermore, each redacted entity in the same category is indexed numerically and recurring entities are consistently labeled with the same index number. An example of an source text and its redacted version is available in Figure \ref{fig:redaction_example}.

This method of indexing is crucial for distinguishing between different entities within the same categories. Without such systematic indexing, the semantic integrity of the sentences can be severely damaged, resulting in a convoluted text that obscures the intended relationships inherent in the source text. This can make human annotation virtually impossible, as the relational context critical for understanding the dialogue can be lost.


\begin{figure}[ht]
\vspace*{-0.25cm}
\begin{center}
\includegraphics[width=7.7cm]{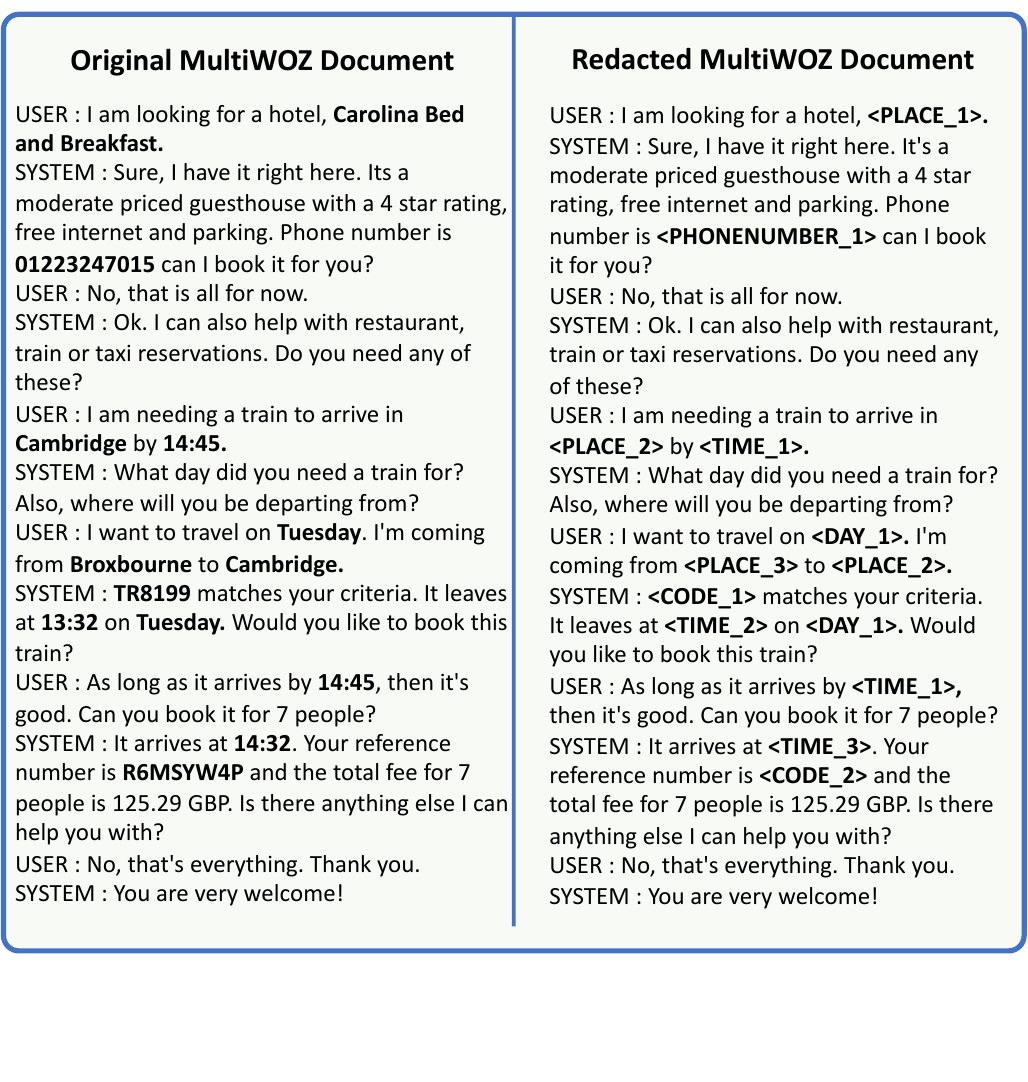}
\end{center}
\vspace*{-1.25cm}
\caption{An example of a redacted MultiWOZ document.}
\label{fig:redaction_example}
\vspace*{-0.5cm}
\end{figure}



\section{Additional Analysis}\label{app:additional_analysis}
This section provides further analysis of the collected human annotations and summary evaluator performance.

\subsection{Challenges in Automated Evaluation}\label{app:IAA_human_machine_mismatch}
 
Table \ref{tab:disagreement_table} shows the ratios of human corrections to LLM's sentence-level binary labels in the fact verification task.
It reveals that human annotators identify the highest frequency of errors for summaries generated by the proprietary LLMs, followed by those generated by the open-source LLMs, and the least for those generated by the non-LLMs. 
This trend highlights that automated faithfulness evaluation is more challenging for summaries generated by recent LLMs compared to those generated by non-LLMs. This can be attributed to 
the fact that factual errors in LLM-generated summaries are more complex and nuanced, often involving subtle misrepresentations that are harder to detect.

\begin{table}[t!]
\footnotesize
\setlength{\tabcolsep}{4.3pt}
\renewcommand{\arraystretch}{1.1}
\begin{center}
\begin{tabular}{ccc}
\toprule
\makecell{Model \\ type}
& \makecell{Human Correction to \\ AI-Claimed Error*
}
& \makecell{Human Correction to \\ AI-Claimed Non-Error**} \\ 
\midrule
\makecell{Non \\ LLM}         
& \makecell{17.83\% \\ (168/942)}            & \makecell{1.79\% \\ (62/3,459)}                \\
\makecell{Open \\ Source \\ LLM} 
& \makecell{20.00\%  \\ (174/870)}            & \makecell{2.66\% \\ (287/10,803)}              \\
\makecell{Prop. \\ LLM} 
& \makecell{22.49\% \\ (56/249)}             & \makecell{2.84\% \\ (229/8,076)}              \\
\midrule
Total           & \makecell{19.31\% \\(398/2,061)}            & \makecell{2.59\% \\ (578/22,338)}               \\ 
\bottomrule
\end{tabular}
\vspace*{-0.2cm}
\caption{Human correction to AI factuality evaluation labels. *Cases where the AI inaccurately flags sentences as factually incorrect, and human annotators correct these errors. **Cases where the AI deems sentences error-free, but human annotators identify factual inaccuracies.}
\label{tab:disagreement_table}
\end{center}
\vspace*{-0.4cm}
\end{table}

\subsection{Detailed Human Annotation Result}\label{app:hallu_full_result}

Tables \ref{tab:faith_score_table_total} --\ref{tab:cpos_score_table_total} presents a comprehensive breakdown of the human annotation results for each domain and model, separately for faithfulness, completeness, conciseness, and the composite score. We present additional domain-level findings.

{\paragraph{Faithfulness Score.} Table \ref{tab:faith_score_table_total} indicates that the faithfulness scores are fairly high across all domains except for non-LLMs. The input type (dialogue vs. non-dialogue) causes more significant variations in faithfulness scores than the domain itself. Specifically, faithfulness scores for dialogue generally range from 61.4\% to 99.2\%, which are lower compared to non-dialogue scores, which range from 74.7\% to 100.0\%.}

{\paragraph{Completeness Score.} Table \ref{tab:comp_score_table_total} reveals that, across all language model categories, completeness scores generally drop significantly in three domains: Report, Medical Literature, and Sci-fi. This suggests that recent summarizers struggle to identify key information in documents characterized by specialized terminologies, as in the Report and Medical Literature domains, or by intricate plots and unique vocabulary, as in the Sci-fi domain.} 

{\paragraph{Conciseness Score.} Table \ref{tab:conc_score_table_total} demonstrates that the conciseness scores are generally high, with the exception of the Sci-fi domain. This finding suggests that verbose summaries generated by recent language models may be attributed to the imaginative content and unconventional plot structures typical of the Sci-fi domain.}

\subsection{The Relationship between Abstractiveness and Faithfulness}\label{app:abstractive_faithfulness}

Table \ref{tab:abs_faith_correlation} shows the Pearson correlation coefficients between the human scores in faithfulness evaluation and abstractiveness scores across the three summarizer categories. The trade-off between abstractiveness and faithfulness exists only for the non-LLMs and open-source LLMs.

\begin{table}[t]
\footnotesize
\setlength{\tabcolsep}{30pt}
\renewcommand{\arraystretch}{1.25}
\begin{center}
\begin{tabular}{cc}
\toprule
Model   type    & $\rho$                           \\ \midrule
Non-LLM         & \cellcolor{pink}-0.24* \\
Open-source LLM & \cellcolor{pink}-0.14* \\
Proprietary LLM & 0.05                           \\ 
\bottomrule
\end{tabular}
\vspace*{-0.2cm}
\caption{Pearson correlation coefficients between human score in faithfulness evaluation and abstractiveness scores. *p-value < 0.05}
\label{tab:abs_faith_correlation}
\end{center}
\vspace*{-0.7cm}
\end{table}

\begin{table*}[!ht]
\begin{center}
\footnotesize
\setlength{\tabcolsep}{3.5pt}
\renewcommand{\arraystretch}{1.25}
\begin{tabular}{
    ccccccccccccc}
\toprule
\multirow{3}{*}{\makecell{Model \\ Type}} &
\multirow{3}{*}{\makecell{Summ. \\ Model}} &
\multicolumn{5}{c}{Non-Dialogue} & \multicolumn{4}{c}{Dialogue} & 
\multirow{3}{*}{\makecell{Avg. \\ Score}} &
\multirow{3}{*}{\makecell{DoS}} \\  
\cmidrule(lr){3-7}
\cmidrule(lr){8-11}
& & 
News & 
Lifestyle & 
Report & 
\makecell{Med lit} & 
Sci-fi & 
\makecell{Daliy life} & 
Booking & 
Interview & 
Meeting & 
\\ 
\midrule

\multirow{2}{*}{\makecell{Non \\ LLM}} &
  BART\textsubscript{large} &
  91.5 &
  83.7 &
  91.0 &
  91.9 &
  85.7 &
  80.3 &
  82.8 &
  81.1 &
  74.7 &
  84.7 &
  82.8 \\
 &
  Flan-T5\textsubscript{large} &
  84.7 &
  96.0 &
  85.3 &
  90.0 &
  74.7 &
  66.0 &
  78.6 &
  89.5 &
  61.4 &
  80.7 &
  65.4 \\
\midrule
\multirow{4}{*}{\makecell{Open \\ Source \\ LLM}} &
  Phi-2 &
  81.8 &
  62.8 &
  \textbf{100.0} &
  75.8 &
  77.6 &
  83.6 &
  90.0 &
  81.1 &
  81.3 &
  81.5 &
  62.8 \\
 &
  Mistral\textsubscript{7B-Inst} &
  94.7 &
  97.2 &
  97.4 &
  98.8 &
  94.7 &
  91.4 &
  97.9 &
  92.8 &
  92.9 &
  95.3 &
  92.7\\
 &
  Llama2\textsubscript{13B-Chat} &
  92.7 &
  \textbf{100.0} &
  96.7 &
  89.7 &
  91.0 &
  91.0 &
  92.5 &
  89.6 &
  83.7 &
  91.9 &
  83.7 \\
 &
  Mixtral\textsubscript{8x7B-Inst} &
  97.4 &
  98.0 &
  99.2 &
  98.2 &
  93.7 &
  98.3 &
  93.9 &
  \textbf{98.9} &
  92.4 &
  96.7 &
  93.1 \\
\midrule
\multirow{3}{*}{\makecell{Prop. \\ LLM}} &
  GPT-3.5\textsubscript{turbo} &
  95.0 &
  96.0 &
  99.6 &
  97.6 &
  94.2 &
  \textbf{99.0} &
  98.0 &
  93.2 &
  94.4 &
  96.3 &
  \textbf{93.6} \\
 &
  GPT-4\textsubscript{turbo} &
  97.3 &
  97.2 &
  98.8 &
  96.3 &
  \textbf{99.3} &
  98.0 &
  100.0 &
  93.2 &
  92.1 &
  96.9 &
  92.1 \\
 &
  Claude2.1 &
  \textbf{99.3} &
  93.3 &
  \textbf{100.0} &
  \textbf{100.0} &
  97.9 &
  97.6 &
  \textbf{99.2} &
  92.3 &
  \textbf{98.6} &
  \textbf{97.6} &
  92.3 \\
\midrule
\multicolumn{2}{c}{Avg. Score in Domain} &
  92.7 &
  91.6 &
  96.5 &
  93.1 &
  89.9 &
  89.5 &
  92.5 &
  90.6 &
  90.2 &
  &
  \\                         
\bottomrule
\end{tabular}
\vspace*{-0.2cm}
\caption{Faithfulness scores of each summarizer across the nine domains. "DoS" indicates domain stability scores for faithfulness.}
\label{tab:faith_score_table_total}
\end{center}
\vspace*{-0.2cm}
\end{table*}

\begin{table*}[!ht]
\begin{center}
\footnotesize
\setlength{\tabcolsep}{3.5pt}
\renewcommand{\arraystretch}{1.25}
\begin{tabular}{
    ccccccccccccc}
\toprule
\multirow{3}{*}{\makecell{Model \\ Type}} &
\multirow{3}{*}{\makecell{Summ. \\ Model}} &
\multicolumn{5}{c}{Non-Dialogue} & \multicolumn{4}{c}{Dialogue} & 
\multirow{3}{*}{\makecell{Avg. \\ Score}} &
\multirow{3}{*}{\makecell{DoS}} \\  
\cmidrule(lr){3-7}
\cmidrule(lr){8-11}
& & 
News & 
Lifestyle & 
Report & 
\makecell{Med lit} & 
Sci-fi & 
\makecell{Daliy life} & 
Booking & 
Interview & 
Meeting & 
\\ 
\midrule

\multirow{2}{*}{\makecell{Non \\ LLM}}  &
  BART\textsubscript{large} &
  49.8 &
  86.0 &
  17.0 &
  23.8 &
  8.0 &
  40.1 &
  55.4 &
  30.8 &
  31.0 &
  38.0 &
  22.0 \\
 &
  Flan-T5\textsubscript{large} &
  43.2 &
  43.7 &
  21.2 &
  19.2 &
  8.4 &
  46.5 &
  42.5 &
  31.1 &
  32.1 &
  38.1 &
  \textbf{61.9} \\
\midrule
\multirow{4}{*}{\makecell{Open \\ Source \\ LLM}} &
  Phi-2 &
  63.9 &
  49.2 &
  19.5 &
  18.2 &
  13.0 &
  46.4 &
  27.4 &
  27.8 &
  23.4 &
  32.1 &
  49.1 \\
 &
  Mistral\textsubscript{7B-Inst} &
  \textbf{71.9} &
  83.5 &
  46.8 &
  44.9 &
  \textbf{37.6} &
  63.3 &
  69.1 &
  55.0 &
  63.8 &
  59.5 &
  54.1\\
 &
  Llama2\textsubscript{13B-chat} &
  50.7 &
  72.6 &
  27.9 &
  48.8 &
  10.8 &
  37.6 &
  43.0 &
  37.7 &
  33.9 &
  40.3 &
  38.2 \\
 &
  Mixtral\textsubscript{8x7B-Inst} &
  69.4 &
  \textbf{92.4} &
  \textbf{47.8} &
  50.7 &
  34.2 &
  64.1 &
  72.7 &
  58.3 &
  62.7 &
  61.4 &
  41.8 \\
\midrule
\multirow{3}{*}{\makecell{Prop. \\ LLM}} &
  GPT-3.5\textsubscript{turbo} &
  53.1 &
  90.4 &
  45.1 &
  \textbf{53.0} &
  29.0 &
  66.0 &
  59.6 &
  47.3 &
  51.3 &
  55.0 &
  38.7 \\
 &
  GPT-4\textsubscript{turbo} &
  66.2 &
  \textbf{92.4} &
  47.6 &
  50.9 &
  36.6 &
  \textbf{76.4} &
  \textbf{76.2} &
  \textbf{61.5} &
  \textbf{67.4} &
  \textbf{63.9} &
  44.2 \\
 &
  Claude2.1 &
  53.4 &
  75.0 &
  26.4 &
  36.1 &
  33.4 &
  67.5 &
  53.4 &
  45.0 &
  44.7 &
  48.3 &
  51.4 \\
\midrule
\multicolumn{2}{c}{Avg. Score in Domain} &
  57.9 &
  76.1 &
  33.3 &
  38.4 &
  23.4 &
  56.4 &
  55.5 &
  43.8 &
  45.7 & \\
\bottomrule
\end{tabular}
\vspace*{-0.2cm}
\caption{Completeness scores of each summarizer across the nine domains.  "DoS" indicates domain stability scores for completeness.}
\label{tab:comp_score_table_total}
\end{center}
\vspace*{-0.2cm}
\end{table*}

\begin{table*}[!ht]
\begin{center}
\footnotesize
\setlength{\tabcolsep}{3.5pt}
\renewcommand{\arraystretch}{1.25}
\begin{tabular}{
    ccccccccccccc}
\toprule
\multirow{3}{*}{\makecell{Model \\ Type}} &
\multirow{3}{*}{\makecell{Summ. \\ Model}} &
\multicolumn{5}{c}{Non-Dialogue} & \multicolumn{4}{c}{Dialogue} & 
\multirow{3}{*}{\makecell{Avg. \\ Score}} &
\multirow{3}{*}{\makecell{DoS}} \\  
\cmidrule(lr){3-7}
\cmidrule(lr){8-11}
& & 
News & 
Lifestyle & 
Report & 
\makecell{Med lit} & 
Sci-fi & 
\makecell{Daliy life} & 
Booking & 
Interview & 
Meeting & 
\\ 
\midrule

\multirow{2}{*}{\makecell{Non \\ LLM}} &
  BART\textsubscript{large} &
  \textbf{90.9} &
  82.3 &
  73.3 &
  81.3 &
  25.8 &
  75.7 &
  83.4 &
  68.5 &
  75.9 &
  73.0 &
  34.9 \\
 &
  Flan-T5\textsubscript{large} &
  89.7 &
  78.0 &
  \textbf{88.3} &
  83.3 &
  35.3 &
  80.3 &
  72.3 &
  74.2 &
  60.6 &
  73.6 &
  46.7 \\
\midrule
\multirow{4}{*}{\makecell{Open \\ Source \\ LLM}} &
  Phi-2 &
  73.8 &
  68.3 &
  40.1 &
  63.4 &
  27.3 &
  75.4 &
  46.3 &
  46.0 &
  53.0 &
  54.8 &
  51.9 \\
 &
  Mistral\textsubscript{7B-Inst}&
  83.6 &
  95.0 &
  78.7 &
  72.8 &
  59.8 &
  78.5 &
  80.7 &
  75.1 &
  75.0 &
  77.7 &
  64.8 \\
 &
  Llama2\textsubscript{13B-chat} &
  90.5 &
  94.7 &
  65.7 &
  85.6 &
  39.4 &
  88.2 &
  71.6 &
  76.8 &
  69.4 &
  75.8 &
  44.7 \\
 &
  Mixtral\textsubscript{8x7B-Inst} &
  82.8 &
  92.5 &
  70.7 &
  70.2 &
  49.8 &
  84.2 &
  70.0 &
  79.5 &
  68.4 &
  74.2 &
  57.3 \\
\midrule
\multirow{3}{*}{\makecell{Prop. \\ LLM}} &
GPT-3.5\textsubscript{turbo} &
  84.9 &
  98.0 &
  87.8 &
  \textbf{91.4} &
  65.1 &
  \textbf{94.1} &
  72.4 &
  79.7 &
  78.7 &
  83.6 &
  67.1 \\
 &
  GPT-4\textsubscript{turbo} &
  85.8 &
  \textbf{95.2} &
  87.8 &
  90.0 &
  \textbf{70.1} &
  87.7 &
  \textbf{86.1} &
  84.5 &
  79.3 &
  \textbf{85.2} &
  74.9 \\
 &
  Claude2.1 &
  85.2 &
  84.0 &
  76.9 &
  85.4 &
  61.7 &
  83.1 &
  68.9 &
  \textbf{85.2} &
  \textbf{86.5} &
  79.7 &
  \textbf{75.3} \\              
\midrule
\multicolumn{2}{c}{Avg. Score in Domain} &
  85.3 &
  87.5 &
  74.4 &
  80.4 &
  48.3 &
  83.0 &
  72.4 &
  74.4 &
  71.9 &
\\
\bottomrule
\end{tabular}
\vspace*{-0.2cm}
\caption{Conciseness scores of each summarizer across the nine domains. "DoS" indicates domain stability scores for conciseness.}
\label{tab:conc_score_table_total}
\end{center}
\vspace*{-0.2cm}
\end{table*}

\begin{table*}[!ht]
\begin{center}
\footnotesize
\setlength{\tabcolsep}{3.5pt}
\renewcommand{\arraystretch}{1.25}
\begin{tabular}{
    ccccccccccccc}
\toprule
\multirow{3}{*}{\makecell{Model \\ Type}} &
\multirow{3}{*}{\makecell{Summ. \\ Model}} &
\multicolumn{5}{c}{Non-Dialogue} & \multicolumn{4}{c}{Dialogue} & 
\multirow{3}{*}{\makecell{Avg. \\ Score}} &
\multirow{3}{*}{\makecell{DoS}} \\  
\cmidrule(lr){3-7}
\cmidrule(lr){8-11}
& & 
News & 
Lifestyle & 
Report & 
\makecell{Med lit} & 
Sci-fi & 
\makecell{Daliy life} & 
Booking & 
Interview & 
Meeting & 
\\ 
\midrule

\multirow{2}{*}{\makecell{Non \\ LLM}}  &
  BART\textsubscript{large} &
  77.4 &
  84.0 &
  60.5 &
  65.6 &
  39.8 &
  65.4 &
  73.9 &
  60.1 &
  60.5 &
  65.2 &
  55.8 \\
 &
  Flan-T5\textsubscript{large} &
  72.5 &
  72.6 &
  65.0 &
  64.2 &
  39.5 &
  64.3 &
  64.5 &
  64.9 &
  51.8 &
  62.1 &
  66.9 \\
\midrule
\multirow{4}{*}{\makecell{Open \\ Source \\ LLM}} &
  Phi-2 &
  73.1 &
  60.1 &
  53.2 &
  52.5 &
  39.3 &
  68.5 &
  54.6 &
  51.6 &
  52.6 &
  56.2 &
  66.1 \\
 &
  Mistral\textsubscript{7B-Inst} &
 \textbf{83.4} &
  91.9 &
  74.3 &
  72.2 &
  64.0 &
  77.8 &
  82.6 &
  74.3 &
  77.3 &
  77.5 &
  72.1 \\
 &
  Llama2\textsubscript{13B-chat} &
  78.0 &
  89.1 &
  63.4 &
  74.7 &
  47.0 &
  72.3 &
  69.0 &
  68.1 &
  62.4 &
  69.3 &
  58.0 \\
 &
  Mixtral\textsubscript{8x7B-Inst} &
 83.2 &
  94.3 &
  72.6 &
  73.0 &
  59.2 &
  82.2 &
  78.9 &
  78.9 &
  74.5 &
  77.4 &
  64.9 \\
\midrule
\multirow{3}{*}{\makecell{Prop. \\ LLM}} &
  GPT-3.5\textsubscript{turbo} &
  77.7 &
  94.8 &
  77.5 &
  \textbf{80.7} &
  53.1 &
  64.4 &
  56.8 &
  57.7 &
  57.7 &
  68.9 &
  58.3 \\
 &
  GPT-4\textsubscript{turbo} &
  83.1 &
  \textbf{94.9} &
  \textbf{78.1} &
  79.0 &
  \textbf{68.7} &
  \textbf{87.4} &
  \textbf{87.5} &
  \textbf{79.7} &
  \textbf{79.6} &
  \textbf{82.0} &
  73.8 \\
 &
  Claude2.1 &
  79.3 &
  84.1 &
  67.8 &
  73.8 &
  64.3 &
  82.7 &
  73.8 &
  74.2 &
  76.6 &
  75.2 &
  \textbf{80.2} \\
\midrule
\multicolumn{2}{c}{
Avg. Score in Domain} &
  78.6 &
  85.1 &
  68.0 &
  70.6 &
  53.6 &
  75.9 &
  73.3 &
  69.3 &
  67.5 & \\
\bottomrule
\end{tabular}
\vspace*{-0.2cm}
\caption{Composite scores of each summarizer across the nine domains. "DoS" indicates domain stability scores for the composite scores.}
\label{tab:cpos_score_table_total}
\end{center}
\vspace*{-0.2cm}
\end{table*}

\subsection{System-level Evaluator Benchmark Result}\label{app:benchmark_system}
\begin{table*}[ht!]
\begin{center}
\scriptsize
\setlength{\tabcolsep}{8pt}
\renewcommand{\arraystretch}{1.1}
\begin{tabular}{ccccccccccc}
\toprule
\multirow{3}{*}{\makecell{Model \\ Type}} &
\multirow{3}{*}{\makecell{Evaluator}} &
\multicolumn{5}{c}{Non-Dialogue} & \multicolumn{4}{c}{Dialogue} \\ 
\cmidrule(lr){3-7} \cmidrule(lr){8-11}
& & 
News & 
Lifestyle & 
Report & 
\makecell{Med lit} & 
Sci-fi & 
\makecell{Daily Life} & 
Booking & 
Interview & 
Meeting \\ 
\midrule
\multirow{2}{*}{QA-Based} &
  UniEval\textsubscript{faith} &
  -0.52 &
  0.74* &
  -0.60 &
  -0.83* &
  -0.68* &
  -0.07 &
  - &
  -0.22 &
  -0.33 \\
 &
  QAFactEval &
  -0.73* &
  0.07 &
  -0.79* &
  0.03 &
  -0.72* &
  0.08 &
  -0.10 &
  -0.28 &
  -0.50 \\
\midrule
\multirow{4}{*}{NLI-Based} &
  SummaC\textsubscript{Conv} &
  -0.73* &
  0.25 &
  -0.44 &
  -0.43 &
  -0.67* &
  -0.38 &
  -0.72* &
  -0.72* &
  -0.63 \\
 &
  AlignScore &
  -0.40 &
  0.24 &
  -0.28 &
  -0.10 &
  0.45 &
  0.80* &
  0.43 &
  -0.35 &
  0.65* \\
 &
  MiniCheck &
  -0.37 &
  0.71* &
  -0.31 &
  -0.47 &
  -0.90* &
  -0.10 &
  0.27 &
  -0.22 &
  -0.05 \\
\midrule
\multirow{3}{*}{LLM-Based} &
  G-Eval\textsubscript{faith} &
  \textbf{0.93*} &
  \textbf{0.78*} &
  \textbf{0.55} &
  0.32 &
  \textbf{0.93*} &
  0.83* &
  0.92* &
  \textbf{0.68*} &
  \textbf{0.80*} \\
 &
  G-Eval+\textsubscript{faith} &
  0.78* &
  0.75* &
  0.37 &
  \textbf{0.68*} &
  0.90* &
  \textbf{0.97*} &
  \textbf{0.93*} &
  0.55 &
  0.72* \\
 &
  FactScore &
  0.13 &
  -0.01 &
  -0.08 &
  -0.10 &
  -0.43 &
  0.40 &
  0.12 &
  0.23 &
  0.38 \\              
\bottomrule
\end{tabular}
\vspace*{-0.2cm}
\caption{
System level rank correlation with human scores in faithfulness evaluation across nine domains (*: p-value < 0.05). For LLM-based methods, G-Eval, G-Eval+, and FactScore are the summary, sentence, and atomic level evaluators.
}
\label{tab:faith_evaluator_score_table_system_level}
\end{center}
\vspace*{-0.5cm}
\end{table*}

\begin{table*}[!ht]
\begin{center}
\scriptsize
\setlength{\tabcolsep}{8pt}
\renewcommand{\arraystretch}{1.1}
\begin{tabular}{ccccccccccc}
\toprule
\multirow{3}{*}{\makecell{Model \\ Type}} &
\multirow{3}{*}{\makecell{Evaluator}} &
\multicolumn{5}{c}{Non-Dialogue} & \multicolumn{4}{c}{Dialogue} \\ 
\cmidrule(lr){3-7} \cmidrule(lr){8-11}
& & 
News & 
Lifestyle & 
Report & 
\makecell{Med lit} & 
Sci-fi & 
\makecell{Daily Life} & 
Booking & 
Interview & 
Meeting \\ 
\midrule
QA-based                        
& UniEval\textsubscript{coh}  & 0.55           & 0.68* & -0.08 & 0.1   & 0.7*  & 0.23  & -     & 0.5   & 0.70*   \\
\midrule
\multirow{2}{*}{NLI-based} 
& Lite\textsuperscript{3}Pyramid & 0.8*           & -0.6  & 0.53  & \textbf{0.77*} & 0.92* & 0.8*  & -     & 0.47  & \textbf{0.87*}  \\
& A\textsuperscript{3}CU    & 0.8*           & 0.32  & 0.65  & 0.57  & \textbf{0.92*} & 0.57  & -     & 0.67* & 0.53  \\
\midrule
\multirow{2}{*}{LLM-based}      
& G-Eval\textsubscript{coh}   & 0.58           & 0.83* & \textbf{0.83*} & 0.73* & 0.72* & 0.8*  & 0.72* & 0.80*  & 0.82* \\                    
& G-Eval+\textsubscript{com}    & \textbf{0.92*} & \textbf{0.88*} & 0.57  & 0.68* & 0.75* & \textbf{0.88*} & \textbf{0.75*} & \textbf{0.92*} & 0.82* \\
\bottomrule
\end{tabular}
\vspace*{-0.2cm}
\caption{
System level rank correlation with human scores in completeness evaluation across nine domains (*: p-value < 0.05). For UniEval and G-Eval, we use their coherence scores since the completeness dimension is not directly supported.
}
\label{tab:comp_evaluator_score_table_system_level}
\end{center}
\vspace*{-0.2cm}
\end{table*}

\begin{table*}[ht!]
\begin{center}
\scriptsize
\setlength{\tabcolsep}{7.8pt}
\renewcommand{\arraystretch}{1.1}
\begin{tabular}{ccccccccccc}
\toprule
\multirow{3}{*}{\makecell{Model \\ Type}} &
\multirow{3}{*}{\makecell{Evaluator}} &
\multicolumn{5}{c}{Non-Dialogue} & \multicolumn{4}{c}{Dialogue} \\ 
\cmidrule(lr){3-7} \cmidrule(lr){8-11}
& & 
News & 
Lifestyle & 
Report & 
\makecell{Med lit} & 
Sci-fi & 
\makecell{Daily Life} & 
Booking & 
Interview & 
Meeting \\ 
\midrule
QA-based  & UniEval\textsubscript{rel} & -0.27 & -0.03 & -0.08        & \textbf{0.53} & 0.77* & 0.13  & - & 0.78*          & 0.47  \\
\midrule
\multirow{2}{*}{NLI-based} &
Lite\textsuperscript{3}Pyramid &
-0.77* &
-0.02 &
0.03 &
0.43 &
0.65 &
0.35 &
- &
0.75* &
0.03 \\
&
  A\textsuperscript{3}CU 
  & -0.5  
  & 0.00    
  & -0.45        
  & 0.00             
  & 0.52  
  & 0.57  
  & - 
  & 0.58           
  & -0.18 \\
\midrule
\multirow{2}{*}{LLM-based} &
  G-Eval\textsubscript{rel} &
  -0.43 &
  0.60 &
  0.05 &
  \textbf{0.53} &
  \textbf{0.88*} &
  \textbf{0.58} &
  0.20 &
  0.58 &
  0.67* \\
 &
  G-Eval+\textsubscript{conc} &
  -0.1 &
  \textbf{0.85*} &
  0.05 &
  \textbf{0.53} &
  \textbf{0.88*} &
  \textbf{0.58} &
  0.02 &
  0.73* &
  \textbf{0.7*} \\
\bottomrule
\end{tabular}
\vspace*{-0.2cm}
\caption{
System level rank correlation with human scores in conciseness evaluation across nine domains (*: p-value < 0.05). For UniEval and G-Eval, we use their relevance scores since the conciseness dimension is not directly supported.
}
\label{tab:conc_evaluator_score_table_system_level}
\end{center}
\vspace*{-0.2cm}
\end{table*}

We report system-level results (See Appendix \ref{app:eval_calculation} for the calculation details) of evaluator performance on our benchmark. Table \ref{tab:faith_evaluator_score_table_system_level}--\ref{tab:conc_evaluator_score_table_system_level} present the correlations between the scores predicted by the automated evaluators and the human scores at the system-level across the three dimensions.

\subsection{{Comparison with Similarity-based Metric} }\label{app:similarity_evaluator}
{
Table \ref{tab:evaluator_simil} shows the summary-level agreement with human scores for conventional similarity-based metrics (i.e., ROUGE-1/2/L \cite{lin2004rouge} and BERTScore \cite{zhang2019bertscore}) across three dimensions (faithfulness, conciseness, and completeness) and composite score (the average of the three dimensions). In all dimensions, similarity-based evaluators show performance comparable to G-Eval+ in a few domains, such as Report, Medical Literature and Sci-fi. However, in general, they exhibit significantly weaker agreement with human scores in all dimensions compared to the LLM-based evaluator.}



\begin{table*}[!t]
\begin{center}
\scriptsize
\setlength{\tabcolsep}{8pt}
\renewcommand{\arraystretch}{1.1}
\begin{tabular}{cc ccccc cccc}
\toprule
\multirow{3}{*}{Dimension} &
\multirow{3}{*}{\makecell{Evaluator}} &
\multicolumn{5}{c}{Non-Dialogue} & \multicolumn{4}{c}{Dialogue} \\ 
\cmidrule(lr){3-7} \cmidrule(lr){8-11}
& &
News & 
Lifestyle & 
Report & 
\makecell{Med lit} & 
Sci-fi & 
\makecell{Daily Life} & 
Booking & 
Interview & 
Meeting \\ 
\midrule
\multirow{5}{*}{Faithfulness} &
ROUGE-1 & 0.12 & 0.08 & ~~0.27* & ~~0.3* & ~~0.18* & 0.04 & - & 0.07 & ~~0.14* \\ &
ROUGE-2 & 0.06 & 0.04 & ~~0.20 & ~~0.20 & ~~0.20 & -0.03 & - & 0.09 & 0.04 \\ &
ROUGE-L & 0.05 & 0.08 & ~0.22* & ~~0.26* & ~~0.16* & 0.03 & - & 0.08 & 0.05 \\ &
BERTScore & 0.10 & 0.10 & ~~0.29* & ~~0.25* & 0.03 & 0.05 & - & ~~0.16* & ~~0.17* \\ &
G-Eval+\textsubscript{faith}$^{\dagger}$ & ~~\textbf{0.63*} & ~~\textbf{0.57*} & ~~\textbf{0.46*} & ~~\textbf{0.55*} & ~~\textbf{0.38*} & ~~\textbf{0.46*} & ~~\textbf{0.59*} & ~~\textbf{0.52*} & ~~\textbf{0.53*} \\  
\midrule
\multirow{5}{*}{Completeness} &
ROUGE-1 & 0.08 & -0.09 & ~~\textbf{0.61*} & ~~0.47* & ~~0.60* & -0.02 & - & -0.03 & ~~0.17* \\ &
ROUGE-2 & 0.10 & 0.01 & ~~0.46* & ~~0.39* & ~~0.54* & 0.01 & - & 0.01 & 0.02 \\ &
ROUGE-L & -0.01 & -0.09 & ~~0.46* & ~~0.38* & ~~0.57* & -0.02 & - & -0.04 & 0.02 \\ & 
BERTScore  & ~~0.18* & -0.10    & ~~0.49*     & ~~0.26*  & ~~0.48*    & -0.07     & -       & 0.07     & 0.11        \\ 
& G-Eval+\textsubscript{com}$^{\dagger}$           & ~~\textbf{0.57*} & ~~\textbf{0.61*}   & ~~0.59*     & ~~\textbf{0.65*}  & ~~\textbf{0.68*}    & ~~\textbf{0.56*}     & ~~\textbf{0.32*}    & ~~\textbf{0.63*}    & ~~\textbf{0.66*}  \\    
\midrule
\multirow{5}{*}{Conciseness} &
ROUGE-1 & ~~0.13 & 0.04 & -0.01 & 0.09 & ~~0.24 & ~~0.27 & - & ~~0.15 & ~~0.19 \\ &
ROUGE-2 & ~~0.14 & 0.04 & 0.00 & 0.08 & ~~0.22 & ~~0.21 & - & 0.10 & 0.09 \\ &
ROUGE-L & ~~\textbf{0.18} & 0.05 & 0.03 & ~~0.15 & ~~0.23 & ~~0.28 & - & ~~0.15 & ~~0.13 \\ &
BERTScore & 0.11 & 0.05 & ~~0.16* & 0.14 & ~~0.21* & ~~0.26* & - & ~~0.24* & ~~0.17* \\ & 
G-Eval+\textsubscript{con}$^{\dagger}$& 0.11 & ~~\textbf{0.39*} & ~~\textbf{0.17*} & ~~\textbf{0.24*} & ~~\textbf{0.44*} & ~~\textbf{0.36*} & 0.02 & ~~\textbf{0.49*} & ~~\textbf{0.45*} \\  
\midrule
\multirow{5}{*}{Composite} &
ROUGE-1    & ~~0.16* & 0.00 & ~~0.39*   & ~~\textbf{0.38*}  & ~~0.42*  & 0.13      & -       & 0.09      & ~0.25**    \\ &
ROUGE-2    & ~~0.16* & 0.03     & ~~0.30*   & ~~0.30*  & ~~0.40*  & 0.09      & -       & 0.09      & 0.06    \\ &
ROUGE-L    & 0.11 & 0.01     & ~~0.32*   & ~~0.35*  & ~~0.40*  & ~~0.14*      & -       & 0.09      & 0.10    \\ &
BERTScore  & ~~0.20* & 0.01    & ~~0.43*  & ~~0.28*              & ~~0.30*       & 0.11      & -       & ~~0.21*     & ~~0.20* \\ &
G-Eval+$^{\dagger}$    & ~~\textbf{0.47*} & ~~\textbf{0.68*}     & ~~\textbf{0.50*}   & -0.04 & ~~\textbf{0.68*}  & ~~\textbf{0.63*}      & ~~\textbf{0.36*}    & ~~\textbf{0.66*}      & ~~\textbf{0.73*}    \\ 
\bottomrule
\end{tabular}
\vspace*{-0.25cm}
\caption{
Agreement with human scores for similarity-based evaluators in evaluations of three dimensions and a composite score (Pearson correlation on summary-level percentage scores). ${\dagger}$ : For comparative purposes, we include the results for G-Eval+, which is identified as the best-performing LLM-based 
evaluator in our main analysis.
}
\label{tab:evaluator_simil}
\end{center}
\vspace*{-0.3cm}
\end{table*}


\section{Details of Manual Annotation}\label{app:annotationdetail}

\subsection{Annotator Qualification Requirements}
For qualification requirements of annotators on MTurk, we select only those with an approval rating above 95\% and at least 1,000 accepted HITs. Also, we administer a qualification test comprising 10 English comprehension questions that simulate the actual annotation tasks. We limit our pool of crowd-sourced workers to those who score 100 on this test and are based in AU, CA, NZ, GB, or US.

\subsection{Annotator Compensation}
Annotators are paid 50\% above the average American minimum wage. We provided a \$25 bonus to annotators who deliver 500 consecutive high-quality annotations.{The total cost of obtaining fine-grained human annotations for the three evaluation dimensions exceeded \$30K for 2,025 summaries.}

\subsection{Attention Check}
\label{sec:attention}

Our human annotation process involves stringent protocols and detailed strategies that filter out low-quality responses. We extensively incorporates novel attention check methods across the three annotation stages -- fact verification, key-fact validation, and key fact alignment -- to eliminate low-quality and insincere submissions. 
For any submissions where annotators fail the attention checks, we reject their submissions. We ban repeated offenders from participating in future tasks.

These methods enable us to selectively collect high-quality annotations while leveraging the cost-effectiveness and time efficiency of crowd-sourcing platforms, which are crucial for scaling our annotation protocol to larger datasets.

\paragraph{Fact Verification Annotation.}
For each source text-summary sentences pair, we include two types of attention checks using a random summary sentence that should always be labeled as factually incorrect. We manually assign a machine label to this sentence as either "factually correct" or "factually incorrect." Annotators should always "disagree" with the "factually correct" machine label and "agree" with the "factually incorrect" machine label. 

\paragraph{Key-Fact Validation Annotation.}
For each key-fact validation annotation task, we include questions with a random, irrelevant key fact that should always be answered as an invalid key fact.

\paragraph{Key-Fact Alignment Annotation.} We introduce two types of attention checks. One type presents a random sentence as if it were a key fact, ensuring that the alignment with all summary sentences is "not aligned." The other type involves inserting fake summary sentences: one is a slight paraphrase of a key fact, which should always be answered as "aligned," and another is a random sentence consistently appearing across all assessments, which should always be classified as "not aligned". 

\end{document}